\documentclass[10pt,twocolumn,letterpaper]{article}

\usepackage[pagenumbers]{cvpr} 

\definecolor{cvprblue}{rgb}{0.21,0.49,0.74}
\usepackage[pagebackref,breaklinks,colorlinks,allcolors=cvprblue]{hyperref}

\def\paperID{7878} 
\def\confName{CVPR}
\def\confYear{2025}

\title{Layered Image Vectorization via Semantic Simplification}

\usepackage{color}
\usepackage{xcolor}
\usepackage{graphicx}
\usepackage{booktabs}
\usepackage{amsmath, amsthm, amsfonts, amssymb}
\usepackage{mathrsfs}
\usepackage{textcomp}
\usepackage{epstopdf}
\usepackage{multirow}
\usepackage{wrapfig}
\usepackage{subfig}
\usepackage{booktabs} 
\usepackage[ruled]{algorithm2e} 
\usepackage{algorithmicx}  
\usepackage{algpseudocode}
\usepackage{ragged2e}
\usepackage[normalem]{ulem}
\usepackage{cleveref}
\usepackage{bm}
\usepackage{makecell}

\SetAlFnt{\small}
\SetAlCapFnt{\small}
\SetAlCapNameFnt{\small}
\SetAlCapHSkip{0pt}

\definecolor{green}{rgb}{0, 0.5, 0}
\definecolor{orange}{rgb}{0.6, 0.3, 0.1}
\definecolor{red}{rgb}{1.0, 0.0, 0.0}
\definecolor{teal}{rgb}{0.0, 0.4, 0.4}
\definecolor{purple}{rgb}{0.65,0,0.65}
\definecolor{saffron}{rgb}{0.95,0.75,0.2}
\definecolor{turquoise}{rgb}{0.0,0.5,0.5}
\definecolor{brown}{rgb}{0.5, 0.16, 0.16}
\definecolor{brickred}{rgb}{.6, .2 .1}
\definecolor{coral}{rgb}{1,0.45,0.33}
\definecolor{newcolor}{rgb}{.8,.349,.1}
\definecolor{ceruleanblue}{rgb}{0.16, 0.32, 0.75}

\newcommand{\lm}[1]{{\color{green} #1}}

\author{
Zhenyu Wang\textsuperscript{1}  \quad Jianxi Huang\textsuperscript{1}  \quad Zhida Sun\textsuperscript{1}  \quad Yuanhao Gong\textsuperscript{1}  \quad Daniel Cohen-Or\textsuperscript{2}  \quad Min Lu\textsuperscript{1}\thanks{Corresponding author, Email: lumin.vis@gmail.com} \\
{\small \textsuperscript{1} Shenzhen University} 
{\small \textsuperscript{2} Tel Aviv University} \\
}

\makeatletter
\def\sf@counterlist{}  
\makeatother

\begin{document}
\maketitle

\begin{abstract}

This work presents a progressive image vectorization technique that reconstructs the raster image as layer-wise vectors from semantic-aligned macro structures to finer details. Our approach introduces a new image simplification method leveraging the feature-average effect in the Score Distillation Sampling mechanism, achieving effective visual abstraction from the detailed to coarse. Guided by the sequence of progressive simplified images, we propose a two-stage vectorization process of structural buildup and visual refinement, constructing the vectors in an organized and manageable manner. The resulting vectors are layered and well-aligned with the target image's explicit and implicit semantic structures. Our method demonstrates high performance across a wide range of images. Comparative analysis with existing vectorization methods highlights our technique's superiority in creating vectors with high visual fidelity, and more importantly, achieving higher semantic alignment and more compact layered representation. 





%

%

\end{abstract}
\section{Introduction}
\label{sec:intro}

Vector graphics represent images at the object level rather than the pixel level, enabling them to be scaled without quality loss\cite{ferraiolo2000scalable}. This object-based structure allows vectors to efficiently store and transmit complex visual content using minimal data, which is ideal for easy editing and integration across various visual design applications. 

Creating effective vector representation, such as SVG, however, often demands considerable artistic effort, as it involves carefully designing shapes and contours to capture an image’s essence. Recently, deep-learning-based methods have been introduced to generate vectors, such as from text descriptions~\cite{jain2023vectorfusion, wu2023iconshop, zhang2024text}. While promising, these methods are constrained by the limitation of pre-trained models, which struggle to produce accurate representation of out-of-domain examples. 

An alternative approach is to generate vectors from raster images, a process known as \textit{image vectorization}. 
Current state-of-the-art vectorization techniques mainly target at visual-faithful reconstruction\cite{ma2022towards, hirschorn2024optimize, zhou2024segmentation}, often produce overly complex and intricate shapes, highlighting the ongoing challenge in achieving a vectorization method that balances visual fidelity with manageability.


\begin{figure}[t]
  \centering    \includegraphics[width=1.\linewidth]{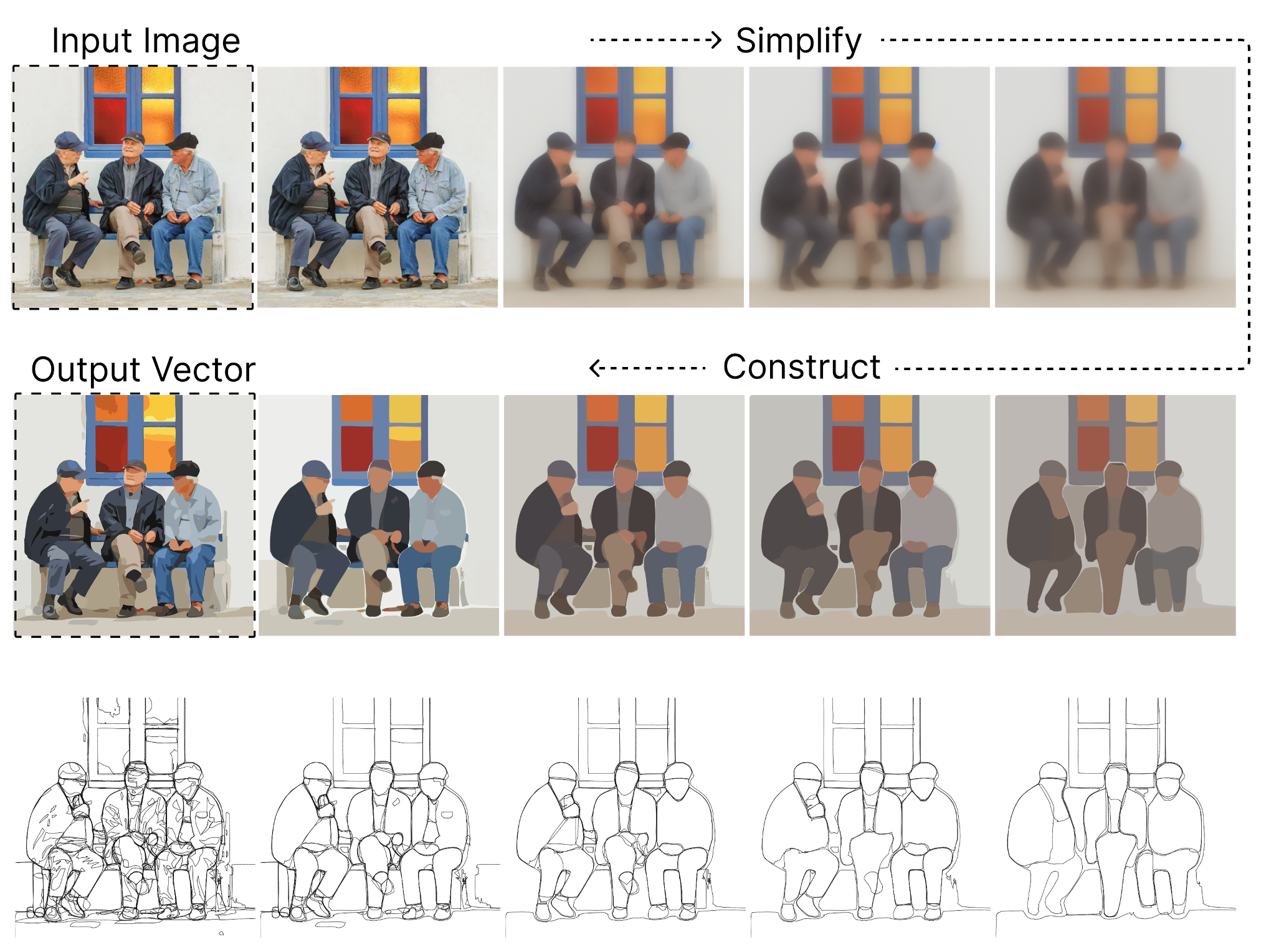}
        \caption{
        Layered vectorization: by generating a sequence of progressive simplified images (top row), our technique reconstructs vectors layer by layer, from macro to finer details (middle row). Our approach maintains the vectors compactly aligned within the boundaries of explicit and implicit semantic objects (bottom row).}
  \label{fig:vec}
\end{figure}

\begin{figure}[t]
  \centering
    \includegraphics[width=1.\linewidth]{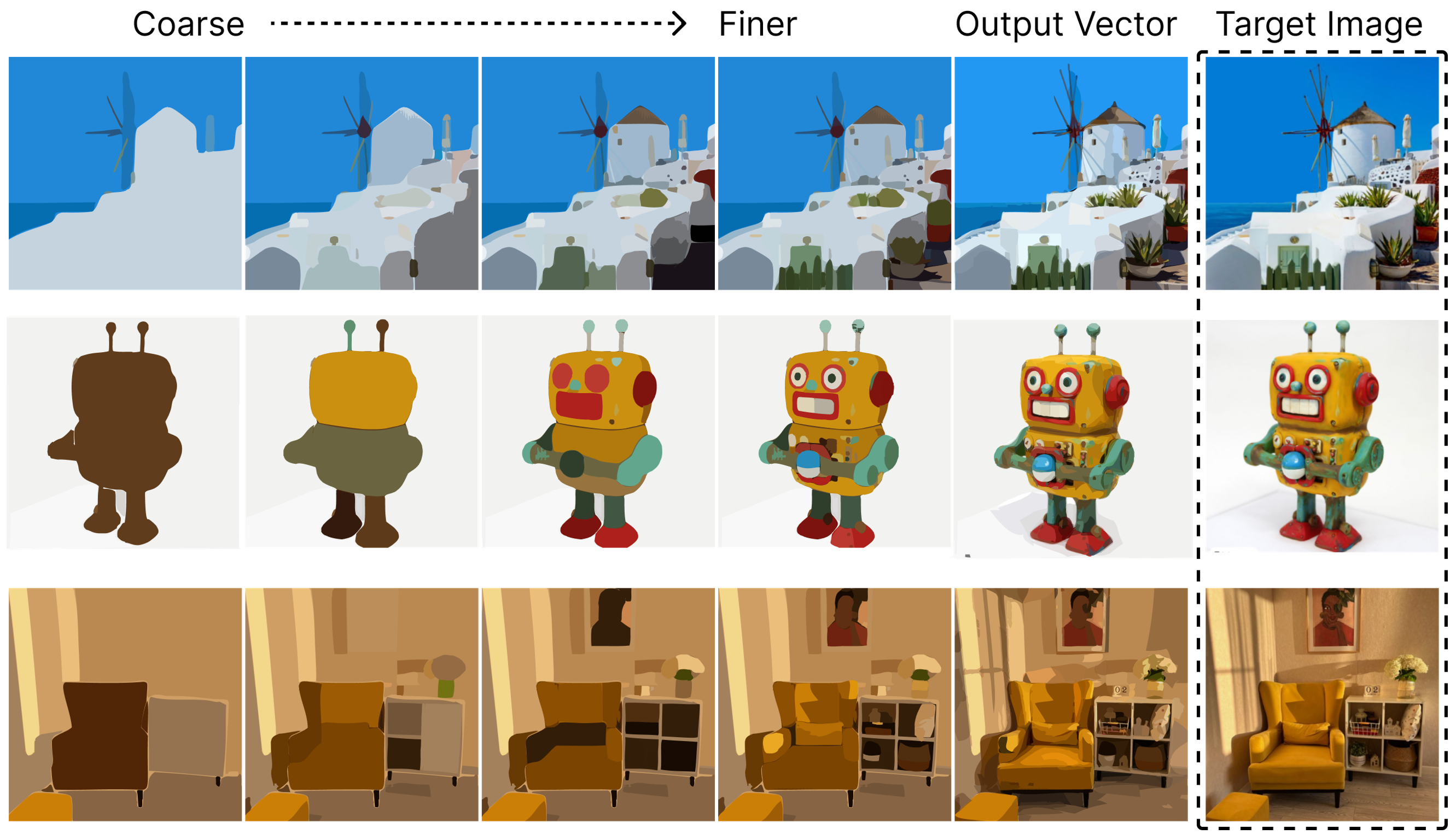}
    \caption{
    Vectors with levels of detail generated with our method: from left to right, vector primitives from macro to finer details are added layer by layer.}
  \label{fig:layered}
\end{figure}

In this paper, we introduce a novel image vectorization approach that progressively creates compact layered vector representations, from macro to fine levels of detail. Unlike existing work that takes the input image as the single target~\cite{ma2022towards, hirschorn2024optimize}, our approach features the process of image semantic simplification, i.e., to generate a sequence of progressively simplified versions for the input image (see the top row in Figure~\ref{fig:vec}). This sequence of simplified images serves as stepping-stone targets, guiding the vector reconstruction with incremental and manageable complexity (see the middle row of Figure~\ref{fig:vec}). Notably, our approach enables the effective deduction of vectors that capture the underlying implicit semantic objects from these simplified abstract images, achieving a layered and semantically aligned vector representation that is highly manageable (Figure~\ref{fig:layered}). 

The key of our approach is a new image simplification technique leveraging Score Distillation Sampling (SDS) in image generation~\cite{poole2022dreamfusion}. By muting the conditioned estimated noise in Classifier-Free Guidance (CFG)~\cite{ho2020denoising}, our method succeeds in harnessing the \textit{feature-average effect}~\cite{hertz2023delta} of SDS for image abstraction, which effectively reduces diverse details while preserving the overall macro structure. Guided by the sequence of progressively simplified images, semantic masks for explicit and implicit objects are detected and layered, based on which vectors are optimized in a two-stage framework. The first stage focuses on building \textit{structure-wise vectors}, optimized towards back-to-front segmented masks by the proposed structure loss. The second stage is to adjust additional vectors to enhance the visual fidelity. 

Our method has been tested on a range of vector-style images (e.g., clipart, emojis), and realistic images. Compared to state-of-the-art methods, our approach demonstrates higher visual fidelity, more compact layer-wise representation, and significant improvement in semantic alignment.

\section{Related work}
\label{sec:formatting}


\paragraph{Image Vectorization} Image vectorization, also known as image tracing, has been a subject of research for decades~\cite{Tian2024_survey,dziuba2023image}. Early work relied on segmentating images into non-overlapping 2D patches, such as triangular~\cite{DEMARET20061604, Battiato2004triagulationrohtua, Swaminarayan2006}, or other irregular shapes~\cite{lecot2006ardeco}. Later research focused on refining the boundaries of these decompositions, introducing curved boundaries to better capture curvilinear features~\cite{xia2009patch, Yang2016Bezigon, liao2012subdivision} and gradients~\cite{Sun2007_Gradient}. 

Another approach to image vectorization tackles it as a curve or polygon fitting problem. This includes methods like diffusion curves, which represent images at the extrema of the gradient1 and then render them through the Poisson equation~\cite{orzan2008diffusion,Xie2014_diffusioncurve, Zhao2018_shapeoptimization}. Other vectorization techniques are particularly suited for non-photorealistic images, such as clip art~\cite{Dominici2020PolyFit, Hoshyari2018_boundary, Favreau2017_photo2clipart, du2023image}, line drawings~\cite{Favreau2016, yan2024deep}, cartoons~\cite{Zhang2009_vectoranimation}, gray-scale manga~\cite{Su2024_mangavec}, and pixel art~\cite{Kopf2011}.

The advent of deep learning has led researchers to approach vectorization through neural networks. A key enabler is the development of differentiable rasterizers~\cite{li2020differentiable}, which bridge the vector and raster domains. Im2Vec~\cite{reddy2021im2vec} employs a variational auto-encoder (VAE) on Fonts and Emoji datasets, to map the input image to a latent space and generate a similar vector. ClipGen~\cite{Shen2022ClipGen} trains an LSTM on clipart vectors of ten categories to optimize vector primitives. Other network-based vectorization methods focus on line-drawing images~\cite{egiazarian2020deep}. Chen et al.~\cite{Chen2023_primitive} leverage a transformer model to assemble vectorizations from simple primitives. Some approaches avoid model learning. LIVE~\cite{ma2022towards} progressively optimize closed cubic Bézier curves to the large difference regions between the rendered SVG and target image. SAMVG~\cite{zhu2024samvg} initializes primitives based on segmented masks from the target raster image. SuperSVG~\cite{hu2024supersvg} trains a model to predict vectors from the superpixel-based segmentation of images. SGLIVE~\cite{zhou2024segmentation} extends the capability of LIVE to support radial gradients via a gradient-aware segmentation. Chen et al.~\cite{Chen2024_texturevec} encapsulate texture in the vector optimization. Hirschorn et al.~\cite{hirschorn2024optimize} introduce an iterative process that adds primitives based on pixel clustering, and removes primitives with low-ranking scores after optimization. 

Our vectorization method is also model-free, leveraging differential rendering. Unlike prior works~\cite{ma2022towards, hirschorn2024optimize, zhu2024samvg, zhou2024segmentation} that optimize towards a single target image, our method generates a sequence of progressively simplified intermediate images as optimization targets. This sequence effectively captures its underlying topology structure from coarse to fine, guiding vectors toward a hierarchical representation of objects and their topologies.

\paragraph{Layer Decomposition} 
Layers are an efficient structure for image manipulation and editing~\cite{Bell2014}. Other image tasks, such as image matting~\cite{smith1996blue} and image reflection separation~\cite{levin2004separating, zhang2018single, hu2023single}, stroke decomposition~\cite{xu2006animating, fu2011animated},  video recoloring~\cite{du2021video, Meka2021video} are also closely related to this topic.

A range of works decomposes images into layered bitmaps, such as single-color layers with varying opacities~\cite{Tan2016Decomposition}, layers with soft colors modeled by a normal distribution~\cite{Aksoy2017soft}, layers with user-specified colors~\cite{koyama2018decomposing}, or color palettes\cite{Wang2019Palette, zhang2021blind, zhang2017palette}. More recently, deep models have been trained to decompose images into transparent layers~\cite{zhang2024transparent}. However, those layers are RGBA semi-transparent layers in raster image format. 

Another bunch of works decomposes images into vectorized layers. Several work vectorize an image into shapes with linear color gradient~\cite{Favreau2017_photo2clipart, Richardt2014rohtua, du2023image}, based on the \textit{alpha compositing}~\cite{porter1984compositing}. Those methods mainly deal with clip art images, which may become prone to errors when applied to natural images. Other work decomposes an image into sequences of brushstrokes to recover the step-by-step painting process~\cite{zou2021stylized, song2024processpainter}. However, these strokes usually lack semantic object-level representation.

\begin{figure}[!htb]
  \centering
    \includegraphics[width=1.\linewidth]{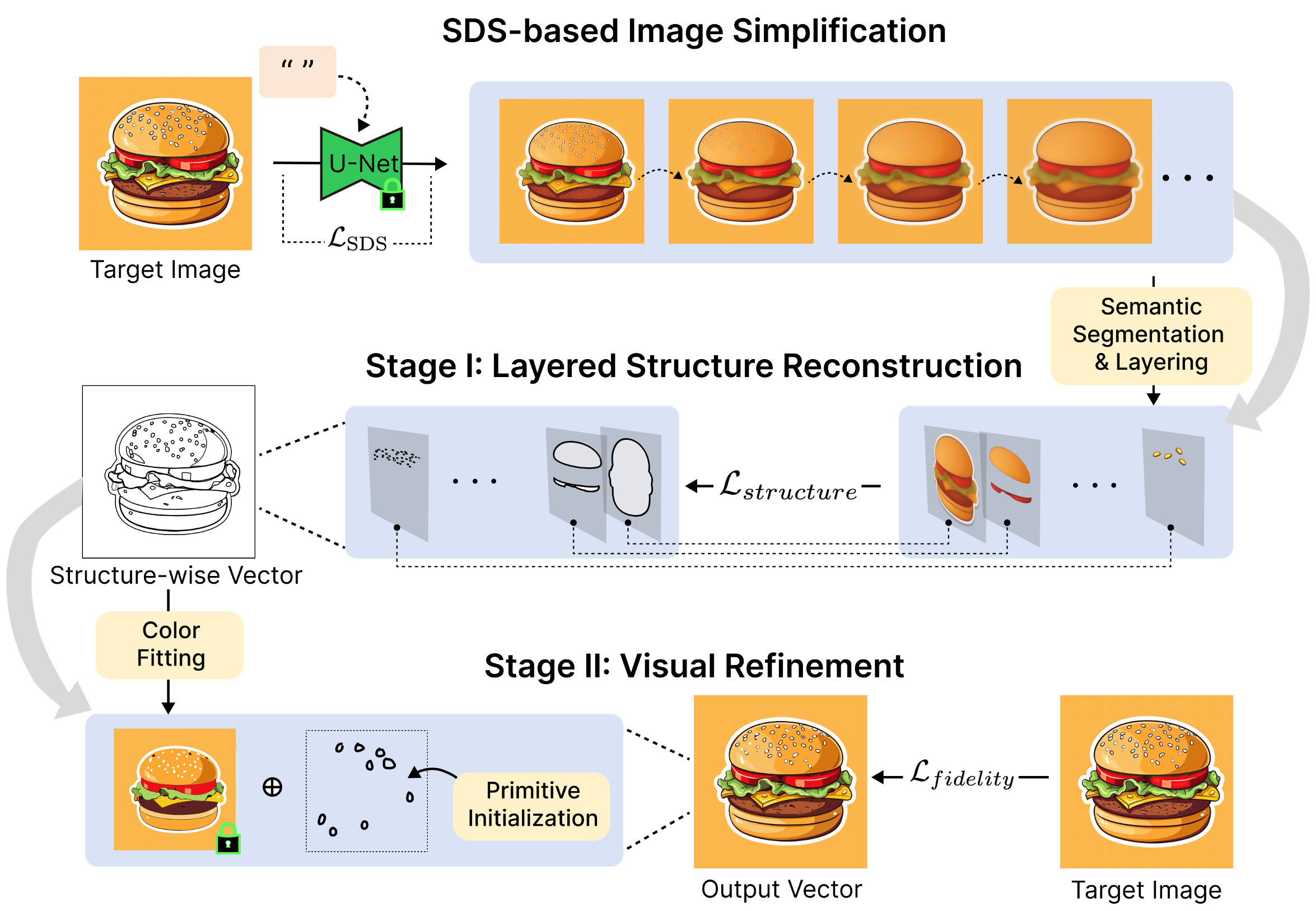}
    \caption{Layered vectorization pipeline: with the input of a target image, its sequence of progressive simplified images is generated using the SDS diffusion model. Vectors are then reconstructed in two stages: structure construction via layer-wise shape optimization to match segmented masks and visual refinement for high fidelity.}
  \label{fig:framework}
\end{figure}

\section{Overview}
\label{sec:overview}

Figure~\ref{fig:framework} shows the pipeline of our method. Taking the target image as input, our process begins with \textit{Progressive Image Simplification} that employs the generative diffusion model based on score distillation sampling~\cite{poole2022dreamfusion}, to generate a sequence of simplified images, 
referred to as the \textit{simplified image sequence}. Taking this simplified image sequence as guidance, our method goes through a two-stage vector reconstruction. 

The first stage involves constructing structure-wise vectors that capture both implicit and explicit semantic structures. To begin, semantic segmentation~\cite{kirillov2023segment} is applied to each image in the simplified sequence, to extract pixels of semantic areas, referred to as \textit{masks}, at various levels of detail. Those segmented masks are sorted and organized into back-to-front layers based on their overlap relationships. Using the raw boundaries of segmented masks as structure-wise vectors can be intricate; therefore, optimizing structure-wise vectors is employed. Specifically, structure-wise vectors are initialized per mask and optimized via differential rendering~\cite{li2020differentiable} with the \textit{layer-wise structure loss}, which measures the shape alignment between structure-wise vectors and their corresponding masks, to ensure the vector representation accurately maintains semantic structures. 



The second stage involves visual refinement. Color fitting is first performed on the structure-wise vectors, serving as the visual basis and frozen. Then taking the rasterized image of frozen structure-wise vectors, we compute its visual differences to the target image. Then visual-wise vector primitives are initialized to regions with large visual differences and optimized toward the target image to minimize the \textit{visual fidelity loss}. During optimization, vector clean-up operations—such as merging and removing redundant vectors—are periodically performed to maintain a neat and efficient vector representation.

\section{Progressive Image Simplification}
\label{sec:sds}






The premise of our vectorization method is to use a sequence of progressively simplified images to guide the optimization. Existing vectorization methods mainly rely on pixel-level analysis of a single target image to decide where to add and optimize vectors, such as large connected areas identified in the target image in LIVE~\cite{ma2022towards} or clusters via DBSCAN in Optimize \& Reduce (O\&R) method~\cite{hirschorn2024optimize}. In contrast, our approach takes the series of simplified images as intermediate targets to optimize. 

\begin{figure}[!htb]
  \centering    \includegraphics[width=1.\linewidth]{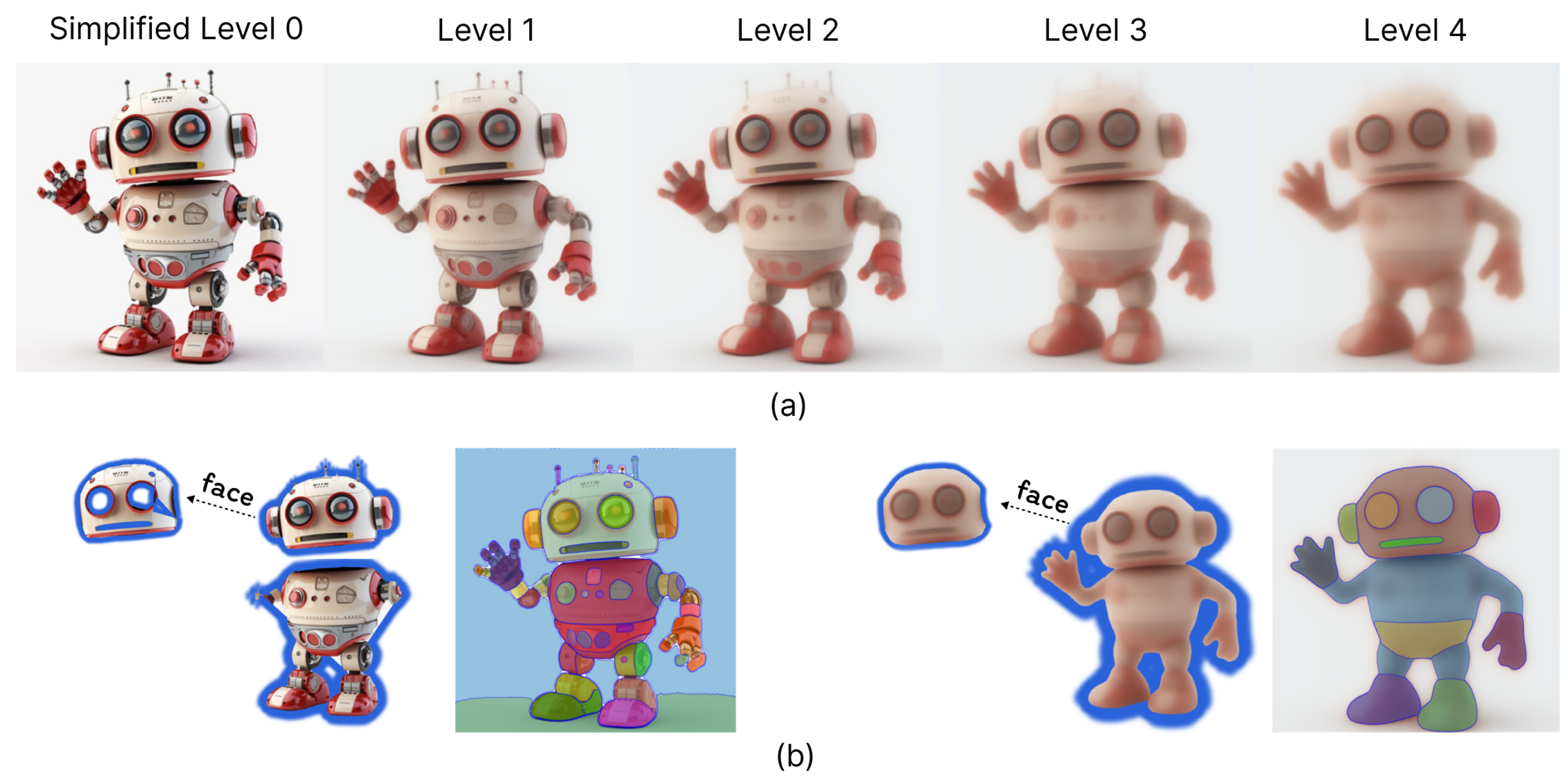}
    \caption{Example of SDS-based image simplification: (a) a sequence of progressively simplified images, with the original image (level 0) on the left; (b) comparison of segmented masks between levels 0 and 4, showing that more simplified image captures more macro structures, such as the 'whole body of the robot' detected in level 4 but not in level 0.}
  \label{fig:sequence_example}
\end{figure}

Figure~\ref{fig:sequence_example}(a) shows an example of the image sequence simplified using our SDS-based method. As shown, the image sequence exhibits varying levels of simplicity, from the original one with many intricate details and textures to a simplified and overall outline on the right.

The intuition behind using simplified images to guide layered vectorization is to prioritize the capture of the overall structure before addressing subtle changes. This approach offers two notable benefits. Firstly, by decomposing the vectorization process into manageable levels, it becomes more tangible and achievable compared to optimizing it as a whole. The incremental improvement from one abstract level to the next is relatively small, allowing for effective optimization at each step. Secondly, the progressive abstraction establishes a macro-to-fine hierarchy, enabling holistic optimization of delicate paths within the image. This approach becomes particularly advantageous when dealing with shapes that exhibit pixel variations due to occlusion, shadows, or textures yet remain integral components of a larger entity. For example, in Figure~\ref{fig:sequence_example}(b), the `robot' is abstracted as a unit shape at the most simplified level, while segmentation of the original target image fails to capture the entire robot shape. Also, its `face' can be represented as a unit shape without holes when the simplified image is applied. 




\subsection{Feature-average Effect in SDS}
\label{subsec:sds}


Our image simplification method takes advantage of the \textit{feature-averaged effect} in SDS~\cite{liang2023luciddreamer, hertz2023delta}. The feature-average effect of SDS can be explained with the gradient of SDS loss, as follows:

\begin{equation}
    \nabla_{\theta} \mathcal{L}_{\text{SDS}}(\theta) \approx \mathbb{E}_{t, \epsilon, c} \left[ \omega(t) \underbrace{\left( \epsilon_{\phi}(\mathbf{x}_t, t, y) - \epsilon \right)}_{\text{SDS update direction}} \frac{\partial g(\theta, c)}{\partial \theta}\right],
\end{equation}

where $g$ is a differentiable generator that maps optimization variables $\theta$ to an image $x = g(\theta)$. For instance, in the original DreamFusion framework~\cite{poole2022dreamfusion}, $g$ is a NeRF volume renderer. In our case, $g(\theta)=\theta$, meaning the optimization variables $\theta$ are the image pixels themselves. The term $\epsilon \sim \mathcal{N}(0, \mathbf{I})$ is the random noise added at timestep $t$, while $\epsilon_{\phi}(\mathbf{x}_t, t, y)$ is predicted noise with given condition $y$. Consequently, $(\epsilon_{\phi}(\mathbf{x}_t, t, y) - \epsilon)$ implies the update direction in the Denoising Diffusion Probabilistic Model (DDPM)~\cite{ho2020denoising}.

\begin{figure}[!htb]
  \centering    
  \includegraphics[width=1.\linewidth]{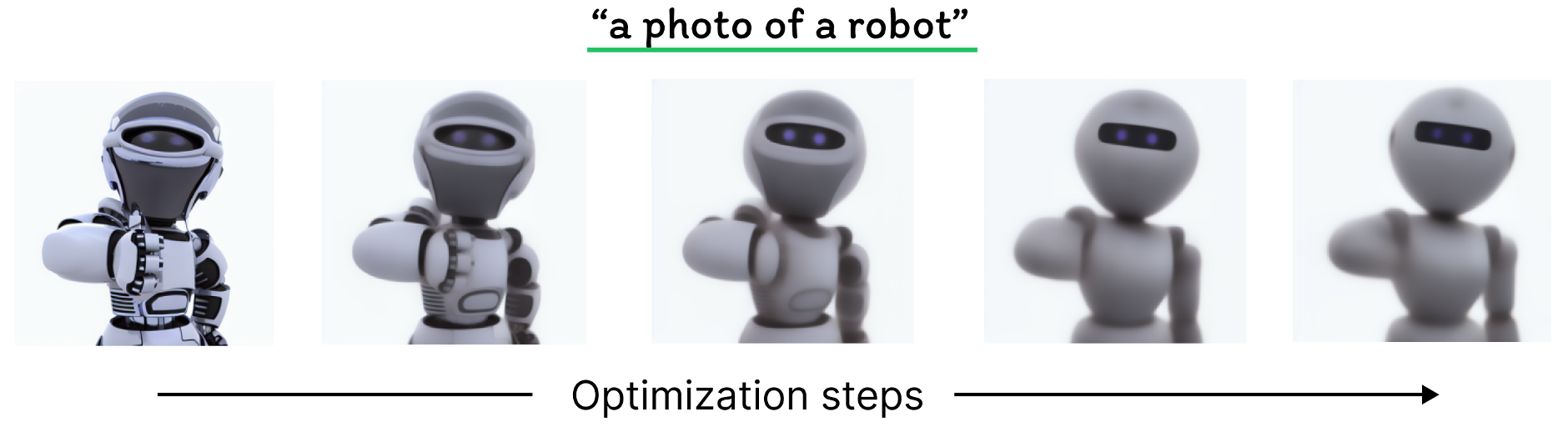}
    \caption{Example of the feature-averaging effect in SDS: as optimization progresses, the `robot' loses fine details, such as `fingers' and `helmet', showing the smoothed and simplified appearance.}
  \label{fig:average_effect}
\end{figure}

It has been observed that the pre-trained DDPM is sensitive to the input, often predicting feature-inconsistent noise, even when conditioning input $y$ remains the same. This causes image pixels $\theta$ to be updated in inconsistent directions, leading to a feature-averaged result. Figure~\ref{fig:average_effect} illustrates the effect using the prompt ``a photo of a robot''. As the number of optimization steps increases, the images progressively lose detailed features, resulting in a more feature-averaged appearance.

\subsection{SDS-based Simplification}

It is crucial to manage the feature-average effect to ensure progressive simplification without significant shape distortion. For instance, as shown in Figure~\ref{fig:average_effect}, while simplification is achieved, it leads to noticeable shape alterations.

In text conditional diffusion models, Classifier-Free Guidance (CFG)~\cite{ho2022classifier} is introduced to combine two predicted noises into one: 

\begin{equation}
    \epsilon_{\phi}^{\omega}(\mathbf{z}_t, y, t) = (1 + \omega) \epsilon_{\phi}(\mathbf{z}_t, y, t) - \omega \epsilon_{\phi}(\mathbf{z}_t, t),
\end{equation}

where $\epsilon_{\phi}(\mathbf{z}_t, y, t)$ is the noise predicted with the conditioned text input and $\epsilon_{\phi}(\mathbf{z}_t, t)$ is that with unconditioned input. The CFG scale controls how closely the conditional prompt should be followed during sampling in DDPM. 

\begin{figure}[!htb]
  \centering    
  \includegraphics[width=1.\linewidth]{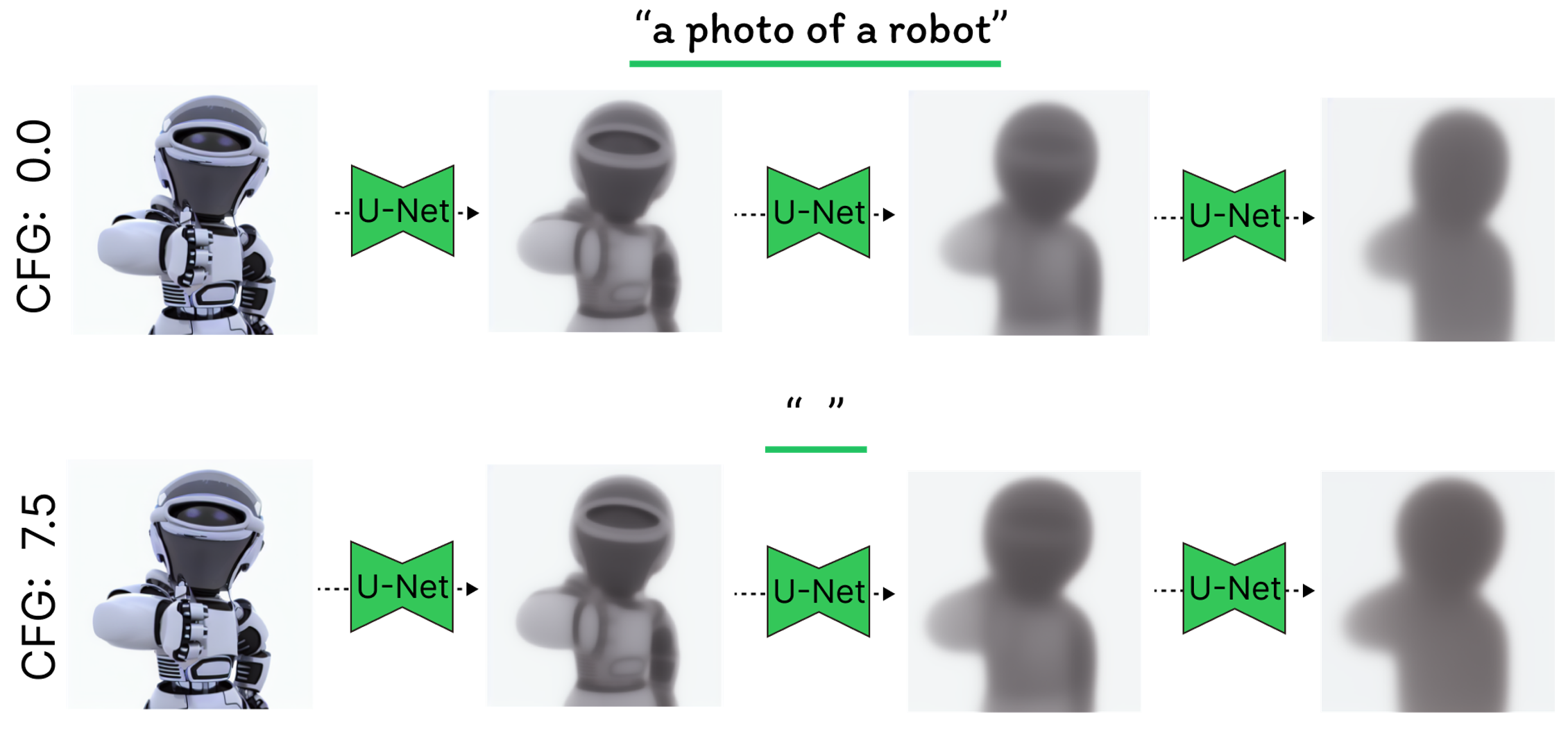}
    \caption{Two approaches for progressive image simplification with macro structure preserved: (top) CFG scale set to zero, (bottom) conditional text prompt set to empty. }
  \label{fig:simplification_con}
\end{figure}

Inspired by CFG, we found that simplification without importing significant shape distortion can be achieved by increasing the portion of unconditional noise. Two possible approaches can be used to achieve this. As shown in Figure~\ref{fig:simplification_con}, one approach is to set the CFG scale to zero. The other approach is to set the conditioned text prompt to empty (i.e., `` ''). Both alternatives are effective; we use the latter one in this work.




\begin{figure}[!htb]
  \centering    
  \includegraphics[width=1.\linewidth]{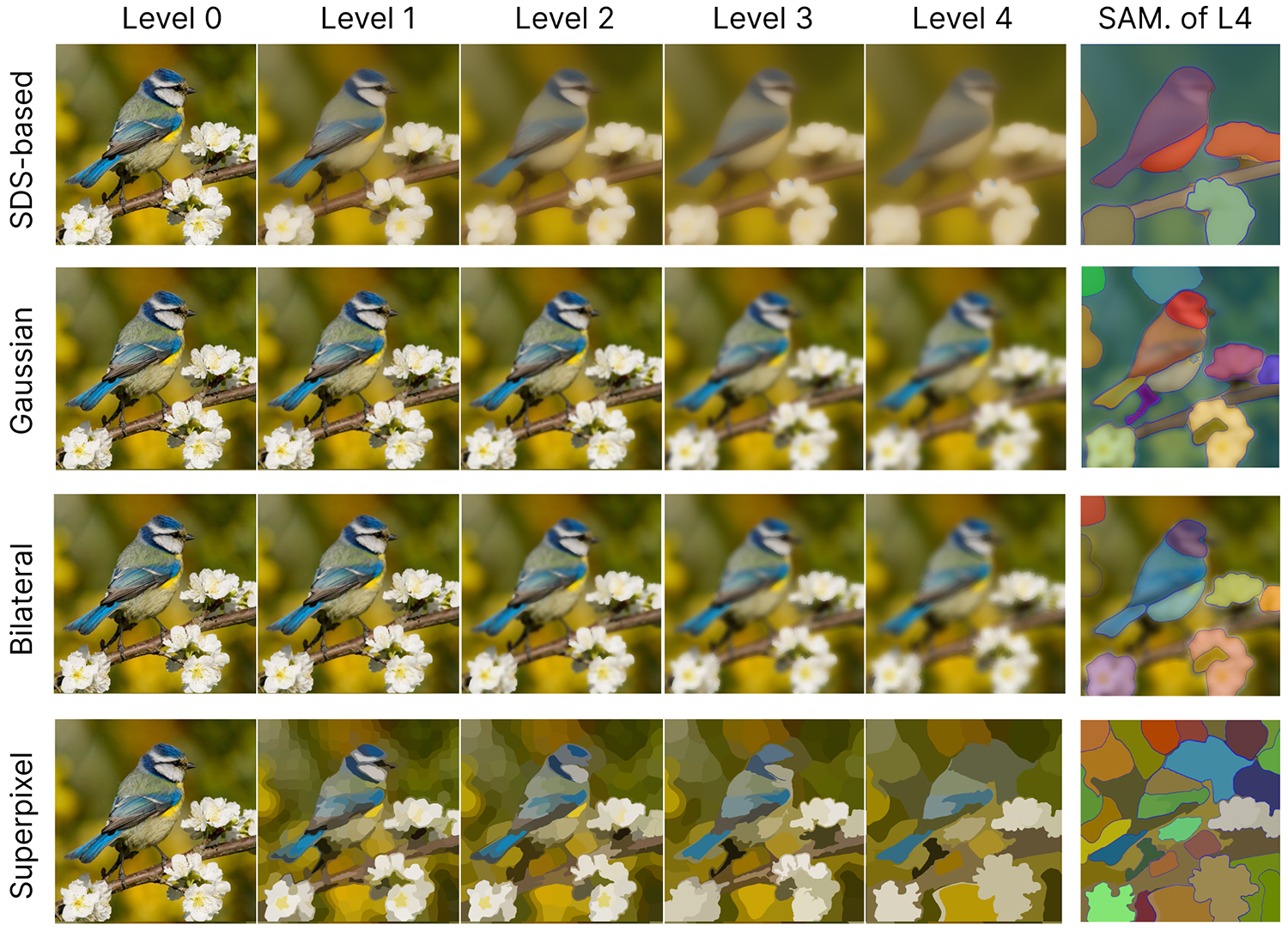}
    \caption{Example of SDS-based image simplification compared to other methods: using the SDS-based method, the macro semantic structures (e.g., `flower') are obtained with smooth boundaries. The rightmost column shows the semantic segmentation of the simplified level 4.}
  \label{fig:abstract_short}
\end{figure}

With the target image as input, a sequence of progressively simplified images is obtained every $N$ optimization steps ($N=20$ in our work).  Figure~\ref{fig:abstract_short} gives an example. Compared to other image simplification methods, e.g., Superpixel~\cite{achanta2012slic}, Bilateral filter~\cite{tomasi1998bilateral}, and Gaussian filter, our SDS-based method effectively abstracts the image while maintaining the macro shapes. Also, the boundaries of detected object masks (the rightmost column of Figure~\ref{fig:abstract_short}) are smoothed and compatible with vector-based graphics.


\section{Vector Reconstruction}

Guided by the sequence of simplified images, vectors are initialized and optimized via differential rendering~\cite{li2020differentiable} in two stages: structural construction and visual refinement. Below, we introduce the key components and loss functions.  

\subsection{Stage I: Structural Construction}
\label{subsec:primitive_init}

\begin{figure}[!htb]
  \centering    
  \includegraphics[width=.90\linewidth]{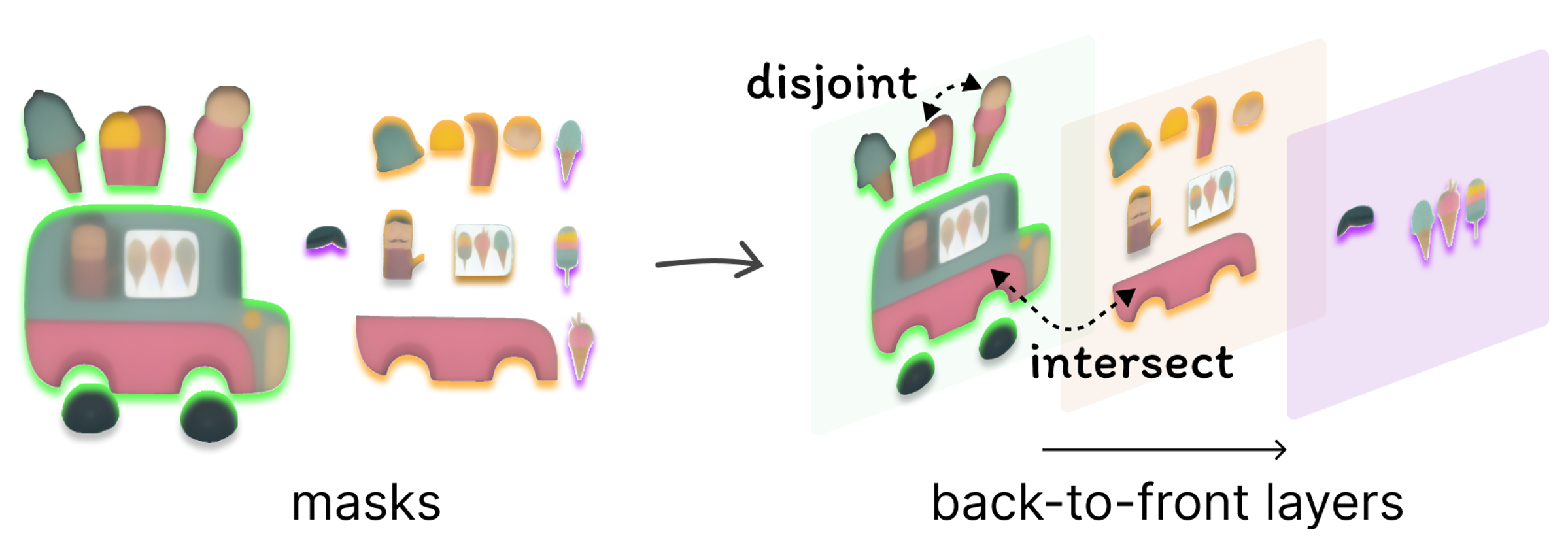}
    \caption{Mask layering: iterating from the most simplified to the least, a mask is only added into the layer from back to front, provided it does not intersect with other masks already in that layer. }
  \label{fig:mask_layer}
\end{figure}

\paragraph{Layering of Segmented Masks} The sequence of simplified images is semantically segmented to detect object masks. Subsequently, the masks are organized into layers, with large masks on the back layer and small masks overlaid on the front. Masks within a layer are not intersected with each other. As illustrated in Figure~\ref{fig:mask_layer}, masks are added progressively from the most simplified image to the least simplified (i.e., the input target), with each mask placed in the furthest back layer where it does not overlap with other masks already positioned on the same layer.

Once the masks are layered, each mask's boundary is 
traced and simplified using the Douglas–Peucker algorithm~\cite{douglas1973algorithms}) to reduce the number of points in the boundary. For each mask, a structure-wise vector is initialized as a closed shape of cubic Bézier curves, with control points set to points in the simplified boundary of the mask. Then structure-wise vectors are rendered by layers and optimized together to minimize the structure loss.

\paragraph{Layer-wise Structure Loss} The structure loss is computing with two components. One component is the layer-wise MSE loss $\mathcal{L}_{\textit{mse}}$, which measures the image differences between each pair of the mask layer $I_{\text{mask}_j}$ and vector layer $I_{\text{vector}_j}$, defined as follows: 

\begin{equation}
\mathcal{L}_{\text{mse}} = \sum_{j=1}^{n}  \| I_{\text{mask}_j} - I_{\text{vector}_j} \|_2^2,
\end{equation}

where $n$ is the number of layers. In the structural construction phase, the focus is on optimizing shape rather than other visual properties such as color. Therefore each pair of mask and corresponding vector are rendered with the same randomly assigned color. 

The second component is the overlap loss $\mathcal{L}_{\textit{overlap}}$, which penalizes the overlap among structure-wise vectors within a layer, defined as follows: 

\begin{equation}
\mathcal{L}_{\text{overlap}} = \sum_{j=1}^{n}  \sum_{p \in {I_{layer_j}}}ReLU(\theta - \alpha(p)),
\end{equation}

where vectors are rendered with the same semi-transparent gray in $I_{layer_j}$, the $\alpha(p)$ is the transparency value of pixel $p$, and $\theta$ is the transparency threshold. Using these components, the structure loss function is a joint loss with weights $w_1 = 1$ and $w_2 = 1e{-8}$: 

\begin{equation}    
\mathcal{L}_{\text{structure}} =  w_1\mathcal{L_{}}_{\text{mse}} + w_2 \mathcal{L}_{\text{overlap}}.
\end{equation}

\subsection{Stage II: Visual Refinement}

\paragraph{Color Fitting} After optimizing the structure-wise vectors to desirable mask shapes, a color fitting is applied to assign a color to each vector. There can be different coloring strategies. In this work, we introduce two. One is to assign the most dominant color from the visible pixels the vector covers in the target image. Another is to fit colors by minimizing the MSE loss between rasterized vectors and the target image. Once the color fitting is accomplished, structure-wise vectors are frozen during subsequent visual refinement. 

To achieve high visual fidelity with the target image, visual-wise vectors are initialized following a strategy similar to LIVE~\cite{ma2022towards}. Vectors are initialized to the top-K largest connected areas with pixel-level differences between the rasterized vectors and the target image, and optimized to minimize visual fidelity loss.


\paragraph{Visual Fidelity Loss} The visual fidelity loss $\mathcal{L}_{\text{fidelity}}$ measures the faithfulness of all vectors to the input image, i.e., how visually similar the rasterized vectors $I_{\text{vector}}$ is to the input target $I_{\text{target}}$. Therefore We define $\mathcal{L}_{\text{fidelity}}$ as their RBG error under $L_2$ norm: 

\begin{equation}
\mathcal{L}_{\text{fidelity}} = \| I_{\text{target}} - I_{\text{vector}} \|_2^2.
\end{equation}

\section{Implementation}

We implemented this method using PyTorch with the Adam optimizer. By default, a sequence of five simplified images (including the original input image) is generated at intervals of 20 SDS iterations. The learning rates for optimizing primitive points and their colors are set to 1.0 and 0.01 respectively. All examples and experiments in this paper were conducted on a system running Ubuntu 20.04.6 LTS, equipped with an Intel Xeon Gold 5320 CPU operating at 2.20 GHz and four NVIDIA A40 GPUs. Each GPU features 48 GB of GDDR6 memory with ECC.

\section{Evaluation}

In this section, we first report the results of the ablation study, and then elaborate on the comparison between our method and four state-of-the-art methods. 

\subsection{Ablation Study}

\begin{figure}[!htb]
  \centering    \includegraphics[width=.9\linewidth]{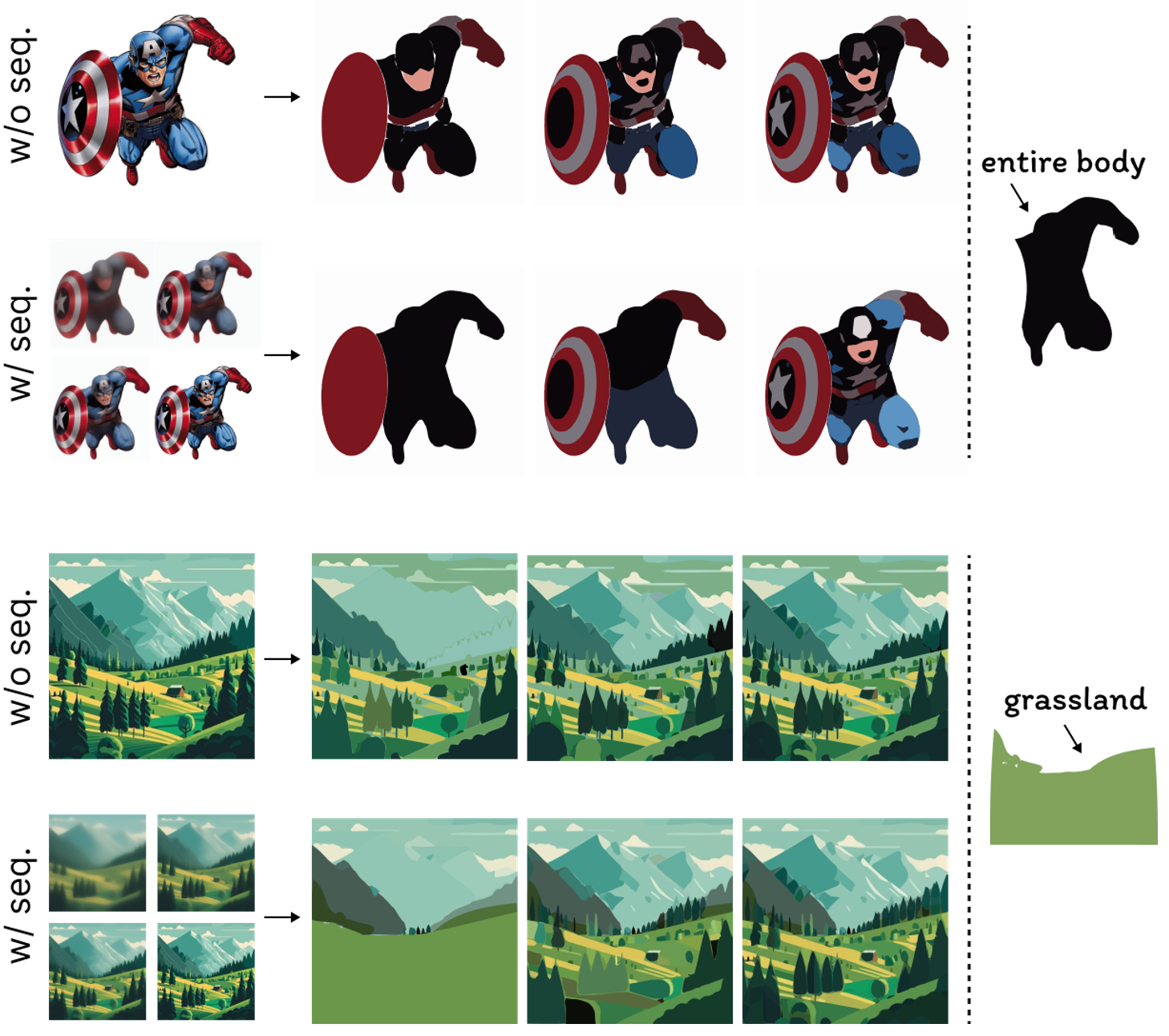}
    \caption{Comparison of structure-wise vectors with and without the simplification sequence: incorporating the simplification sequence creates more implicit semantic vectors, e.g., the `entire body of Captain America', and `grassland'.}
  \label{fig:diff_seq}
\end{figure}

\paragraph{Ablation on the Guide of Simplified Image Sequence} We investigated the impact of incorporating \textit{a sequence of simplified images} in the vectorization process, compared to an ablated version that relies solely on a single input image, i.e., \textit{without the sequence.} As shown in Figure~\ref{fig:diff_seq}, using the sequence of simplified images as the intermediate targets enables our approach to create more implicit semantic vectors, such as the `entire body of Captain America', and the `grassland', which are missed when the sequence is not used. These richer layers allow for fine-grained management of vector elements.

\begin{figure}[!htb]
  \centering    \includegraphics[width=1.\linewidth]{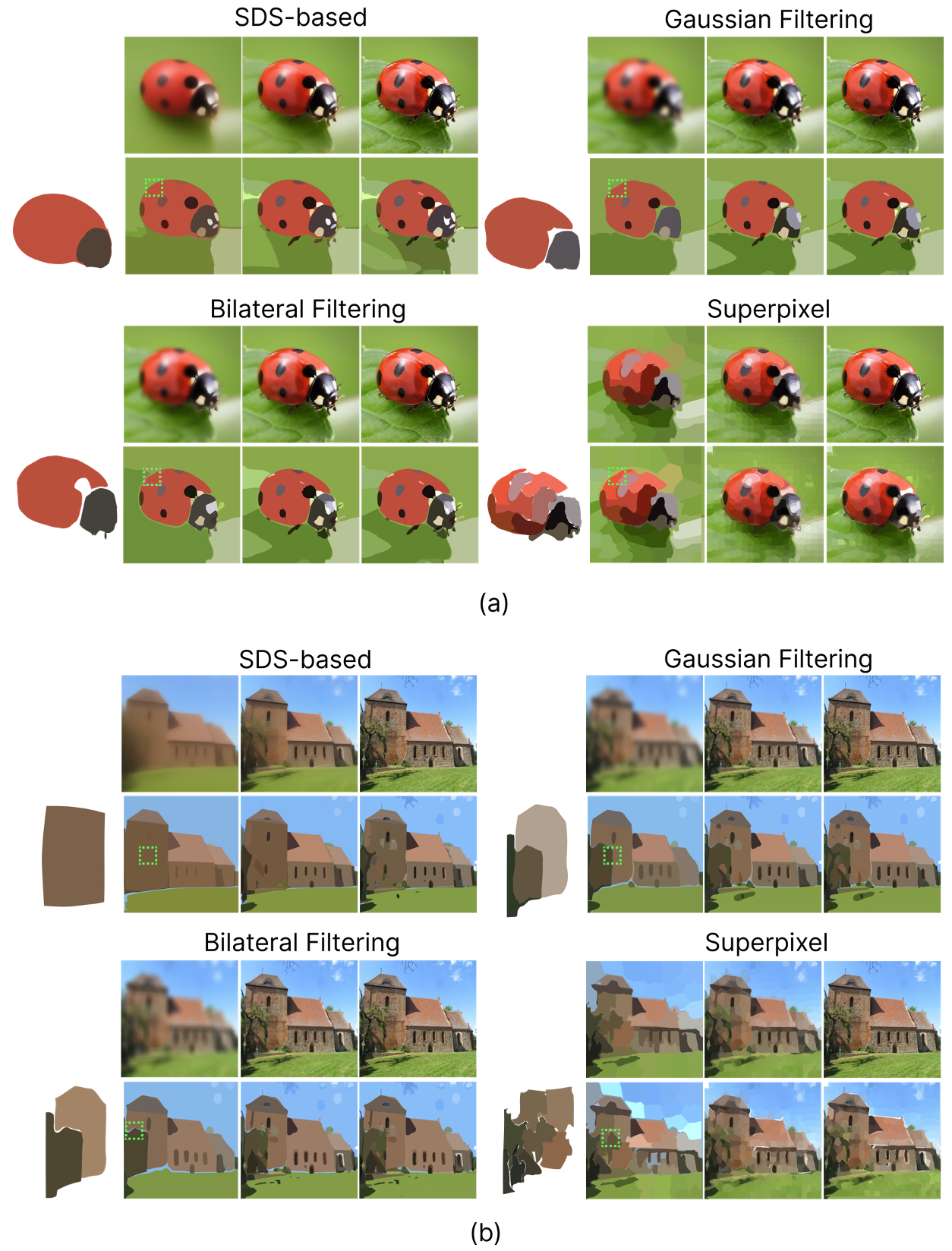}
    \caption{Comparison of structure-wise vectors between SDS-based method and three conventional image simplification methods: (a) SDS-based method retains the round boundary of the `ladybird', (b) and semantically recovers the `front wall of the house'.}
  \label{fig:diff_sim}
\end{figure}

\paragraph{Ablation on SDS-based Image Simplification} We evaluated the effectiveness of \textit{SDS-based simplification} against three conventional image simplification methods: Bilateral filtering, Gaussian filtering, and Superpixel-based Simplification. As shown in Figure~\ref{fig:diff_sim}(a), the SDS-based method preserves clear boundaries, such as the round shape of `the ladybird', more effectively than the other three methods. Also, the SDS-based method smartly removes less featured elements, such as the `trees in front of the house', and recovers the occluded parts of the house in Figure~\ref{fig:diff_sim}(b). The complete constructed vector sequences are provided in the supplementary materials.

\subsection{Comparison Experiment}
\label{subsec:comparison}

We compared our method to existing methods, including DiffVG~\cite{li2020differentiable}, LIVE~\cite{ma2022towards}, O\&R~\cite{hirschorn2024optimize}, and SGLIVE~\cite{zhou2024segmentation}. To make the comparison fair, we use the same visual primitives for all five methods. More details on the experiment set-up and results are provided in the supplementary materials. 



\begin{figure}[!htb]
  \centering    \includegraphics[width=1.\linewidth]{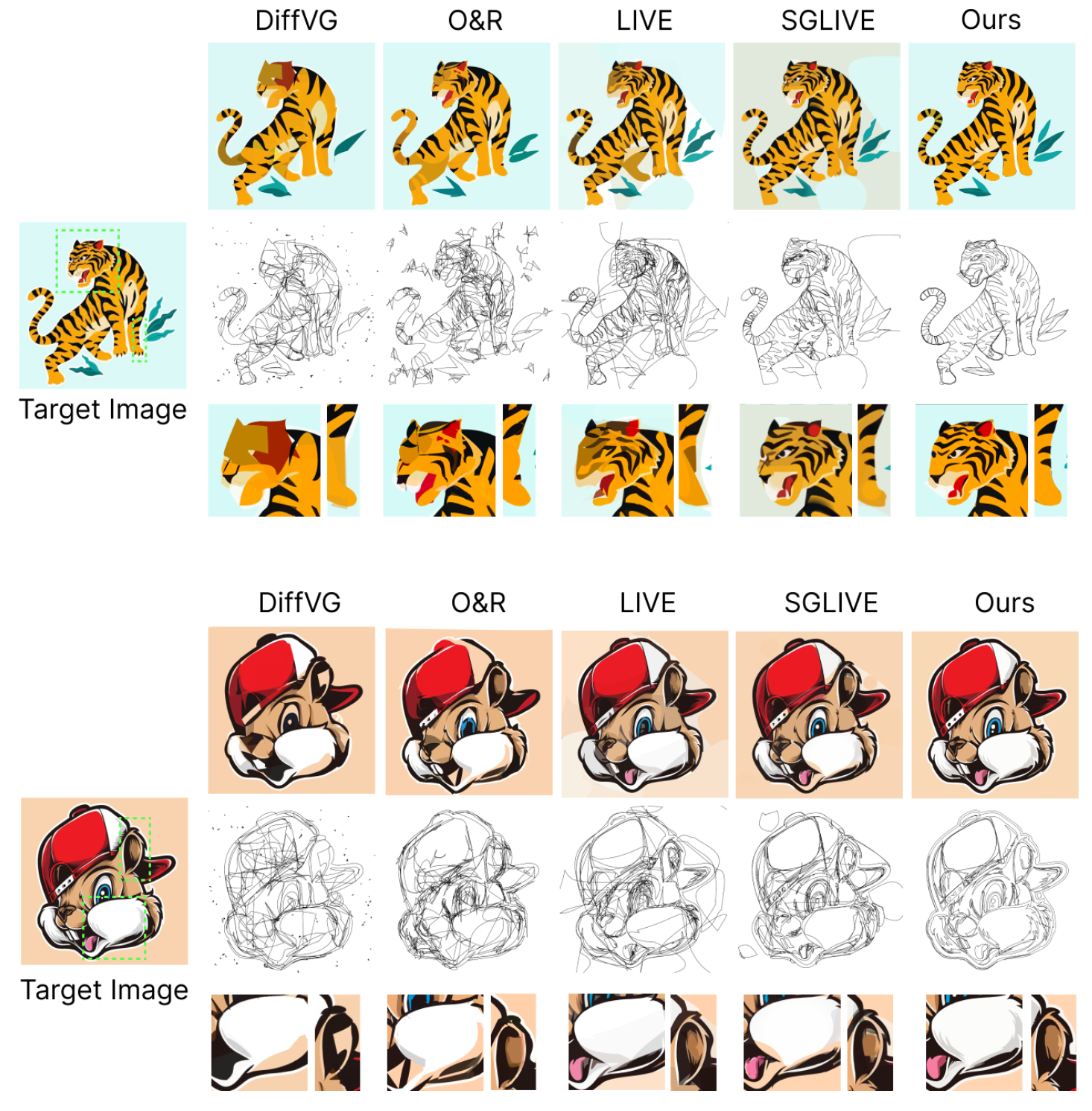}
    \caption{Qualitative reconstruction comparison: both examples are vectorized with 128 vector primitives. Our method reconstructs more faithful, clean, and semantic-aligned vectors.}
  \label{fig:result_mse}
\end{figure}

\paragraph{Visual Quality}  We examined the visual rendering fidelity of the generated vectors, specifically how closely they resemble the original input image. Figure~\ref{fig:result_mse} shows the visual comparison with the four state-of-the-art methods. Clearly, our method demonstrates a more faithful reconstruction with clean, compact, and semantically aligned boundaries, while the other four methods exhibit artifacts in colors and shapes.


\begin{figure}[!htb]
  \centering    \includegraphics[width=1.\linewidth]{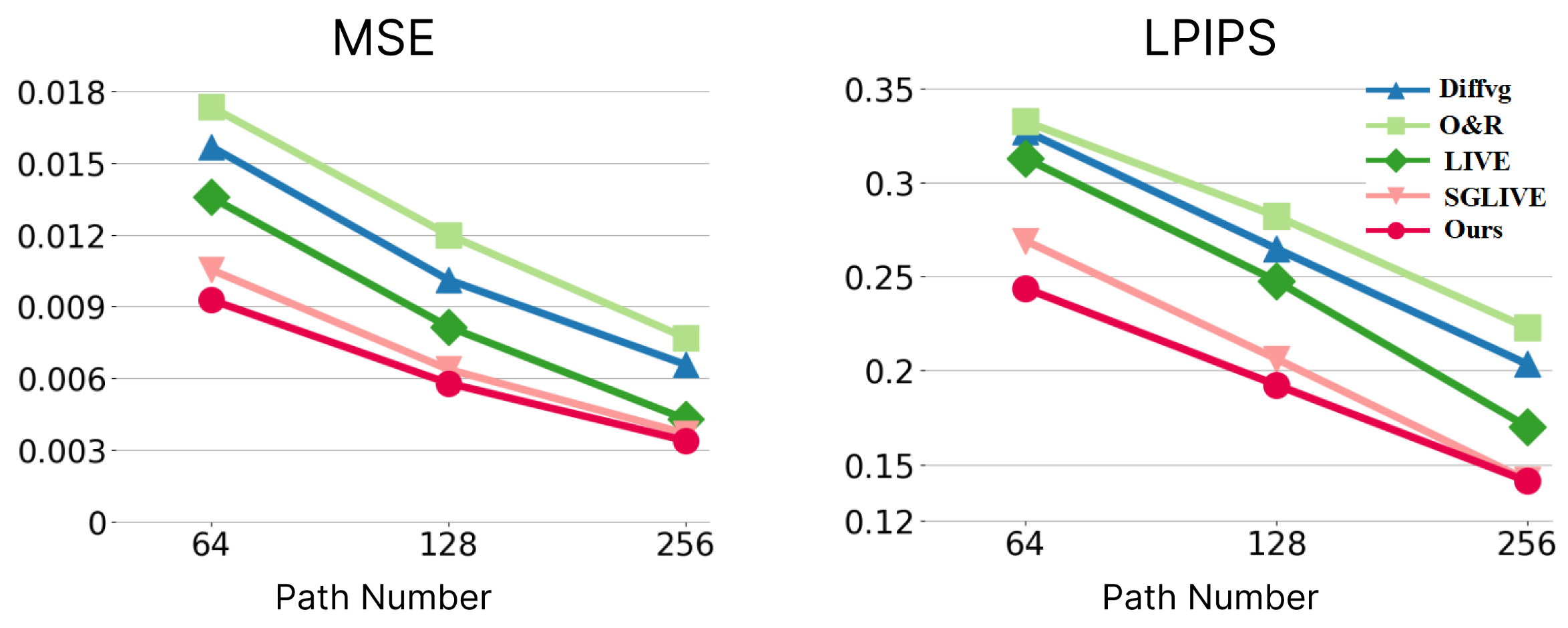}
    \caption{MSE and LPIPS comparison: our method reconstructs more faithful vectors across different numbers of vector primitives.}
  \label{fig:mse_plot}
\end{figure}

To quantify this, we calculated the pixel MSE and LPIPS computed based on VGG~\cite{Simonyan2015vgg} between the rasterized image of vectors and the original image. We collected a testing dataset of 100 images, including realistic photos, clipart images, emojis. Figure~\ref{fig:mse_plot} shows the result on this testing dataset. Our method reconstructs with lower MSE and LPIPS than the other four methods. 


\paragraph{Layer-wise Representation} 



To quantify this, we introduce the metric \textit{Vector Compactness (VeC)}  to assess how well the primitives are contained within the semantic object boundaries. Given a semantic mask (e.g., the area of pixels segmented as `butterfly'), VeC is defined 
as the ratio of vectors highly contained within the mask (i.e., exceeding 85\% area overlap) to the total number of vectors interacting with the mask.

Table~\ref{table:vec}
reports the average VeC of all images in our testing dataset. For each image, we sampled four masks randomly from its semantic segmentation. As can be seen, our method maintains significantly higher compactness of primitives compared to the other four methods, with approximately 73.8\%. 

\begin{table}[!h]
\small
\centering
\caption{Comparison of the average VeC and standard deviation of the 100 testing images.}
\label{fig:compare_vec}
\begin{tabular}{c|c c c c c}
\Xhline{1pt}  
     VeC (\%) & DiffVG & LIVE  & O\&R  & SGLIVE & Ours   \\
        \hline
Avg. & 41.9    & 43.4 & 39.9 & 65.9    & \textbf{73.8}$\uparrow$ \\
Std.  & 15.1   & 17.4 & 20.4 & 18.5    & 11.9
\\
\Xhline{1pt}  
\end{tabular}
\label{table:vec}
\end{table}

\begin{figure}[!htb]
  \centering \includegraphics[width=1.0\linewidth]{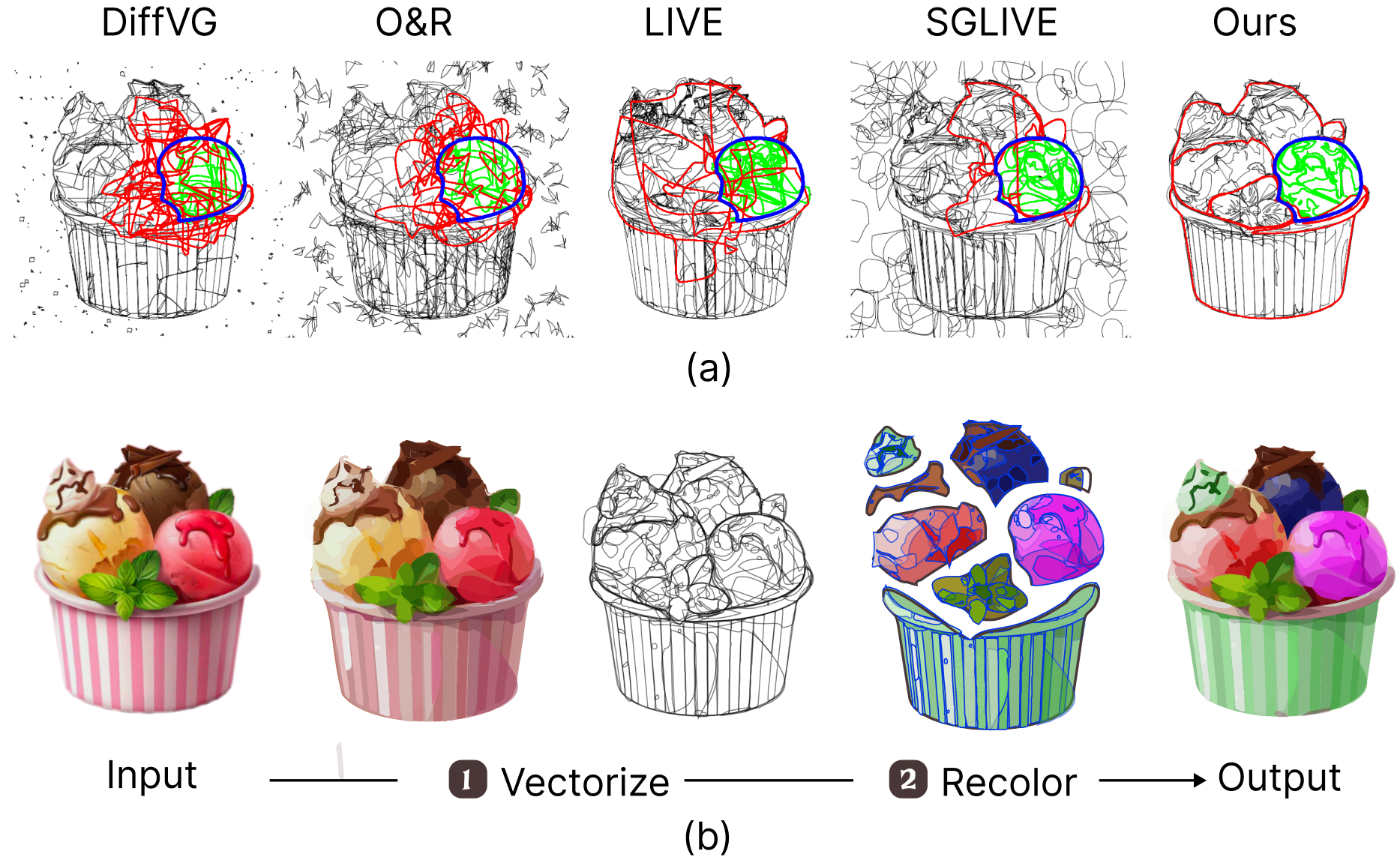}
    \caption{Layered representation and editing: (a) looking at the mask of `ice cream ball' (highlighted in blue), vectors highly contained in the ice cream are colored in green, those intersected but not contained are in red. (b) our compact layer-wise vector representation facilitates recoloring.}
  \label{fig:quick_recolor}
    \captionsetup[figure]{skip=0pt} 
\end{figure}

As visually evident in Figure~\ref{fig:quick_recolor}(a), vector primitives in methods like LIVE and O\&R tend to exhibit scattering and inter-region intersections (highlighted in red). Our method demonstrates superior performance by retaining a large portion of the vector primitives within coherent semantic structures (green). This compact layered construction facilitates vector editing tasks, such as image recoloring. For example, in Figure~\ref{fig:quick_recolor}(b), upper-layer primitives can be easily selected using the underlying layer structures and recolored by applying a specified hue shift. Our method effectively preserves the details of texture and lighting, even after recoloring. 

\begin{figure}[t]
  \centering
    \includegraphics[width=1.\linewidth]{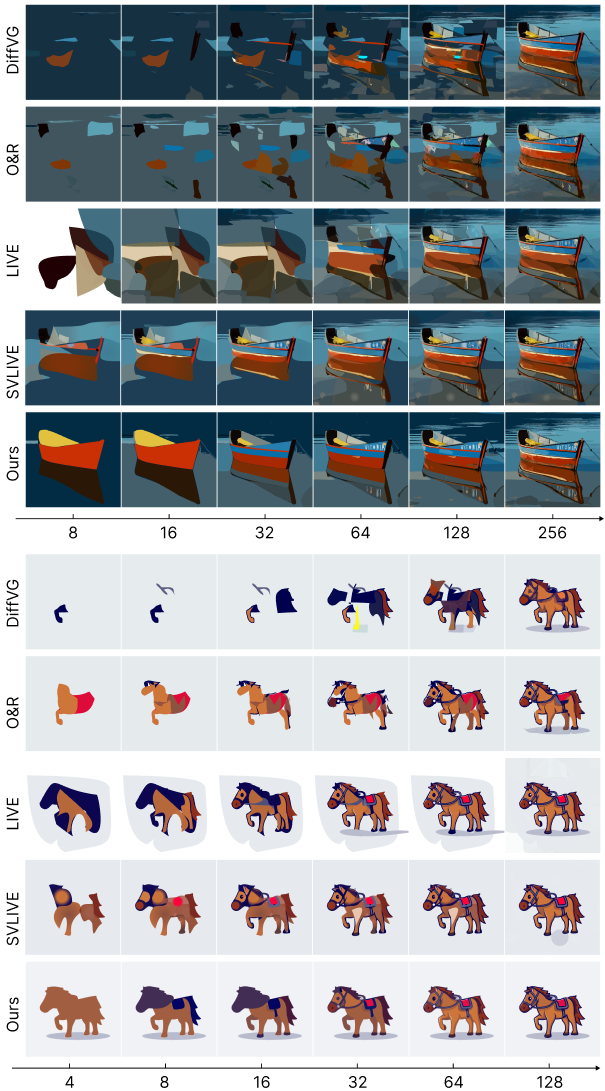}
    \caption{
    Vector layers generated by LIVE~\cite{ma2022towards}, DiffVG~\cite{li2020differentiable}, O\&R~\cite{hirschorn2024optimize}, SGLIVE~\cite{zhu2024samvg}, and our method.}
  \label{fig:exp_layer}
\end{figure}

\paragraph{Semantic Alignment} We examined how well the underlying vector primitives align with meaningful structures in the image. Figure~\ref{fig:exp_layer} shows the two examples (128 paths in `horse' and 256 in `boat'), comparing the sequence of intermediate accumulated vectors from back to front layers. As shown, our method outperforms others, generating primitives that progress from macro to fine detail and align more accurately with the underlying semantic structures. 

\begin{figure}[t]
  \centering
    \includegraphics[width=1.\linewidth]{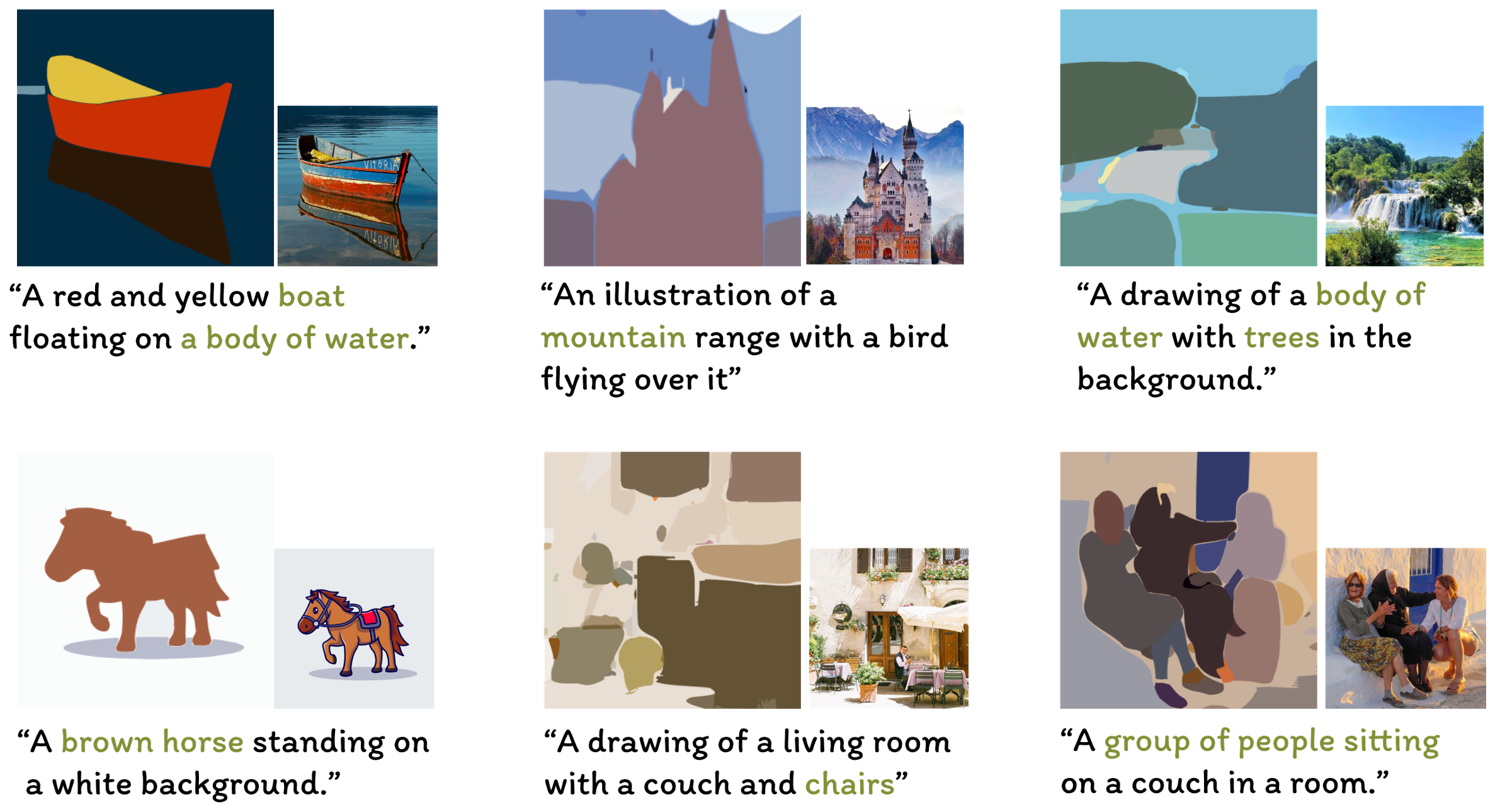}
    \caption{
    Captioning of macro structural vectors generated by our vectorization method: for each example, the caption of the coarse image is generated by Florence-2 model~\cite{florence}.}
  \label{fig:caption}
\end{figure}

Figure~\ref{fig:caption} shows some captions generated for coarse vector layers of our vectorization method. Florence-2 model is applied to infer captions from images. It can be seen the descriptive text generated from coarse layers aligns well with the contents in the original target image.

\begin{figure}[!htb]
  \centering    \includegraphics[width=.6\linewidth]{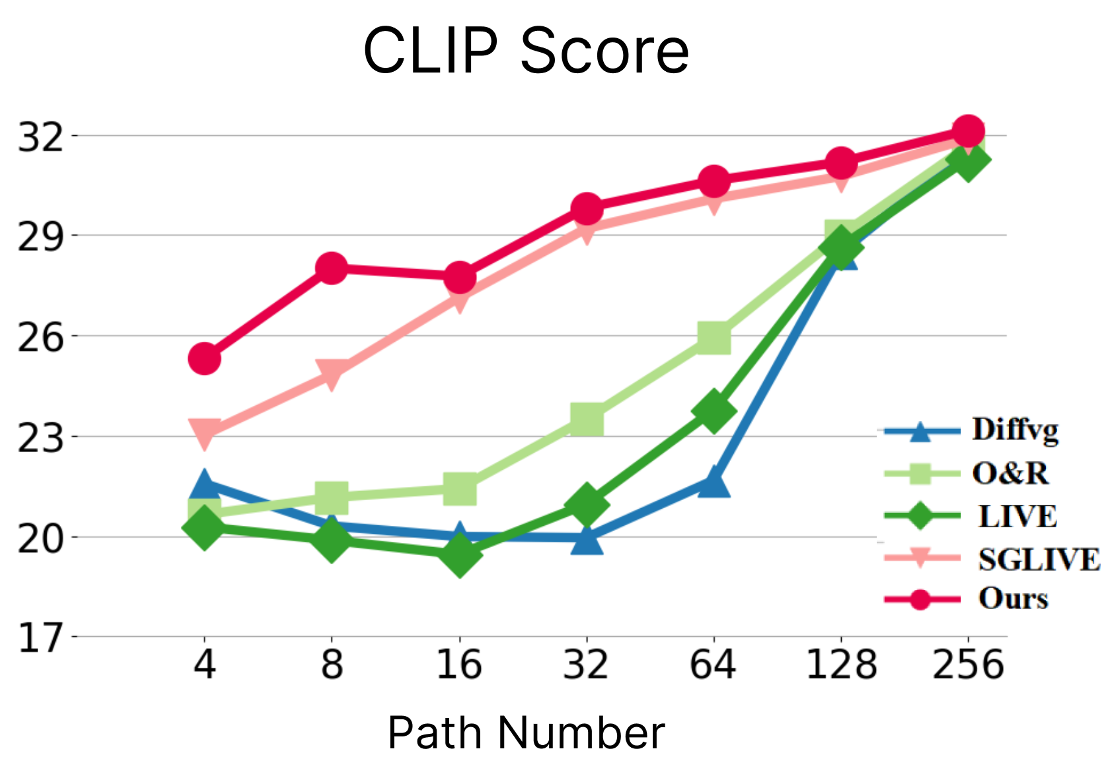}
    \caption{Comparison of CLIP similarity score: our method manages to preserve key semantic features of the original image better than other methods, especially with a few vectors.}
  \label{fig:clip}
\end{figure}

To quantify this, we used the CLIP score~\cite{radford2021learning} to measure the semantic similarity between the generated caption of the input target image and the rasterized SVG images. As shown in Figure~\ref{fig:clip}, at every sampled path number, our method produces vectorized images that retain a closer match to the original image’s content, particularly at the beginning stages where there are only a few vector paths. 


\section{Conclusion}

In this work, we present a novel image vectorization technique that utilizes a sequence of progressively simplified images to guide vector reconstruction. Our approach introduces an SDS-based image simplification method that achieves effective visual abstraction. Through a two-stage vector reconstruction process, our approach emphasizes both visual fidelity and structural manageability, producing a layered vector representation that captures the target image’s structure from macro to fine details. Our method demonstrates superior performance in multiple areas, including visual fidelity, layered representation, and semantic alignment of vectors with underlying structures, greatly enhancing usability for further editing and modification.

{
    \small
    \bibliographystyle{ieeenat_fullname}
    \bibliography{main}
}

\clearpage
\setcounter{page}{1}
\maketitlesupplementary

\section{Additional Ablations}

\begin{figure}[!htb]
  \centering    \includegraphics[width=1.\linewidth]{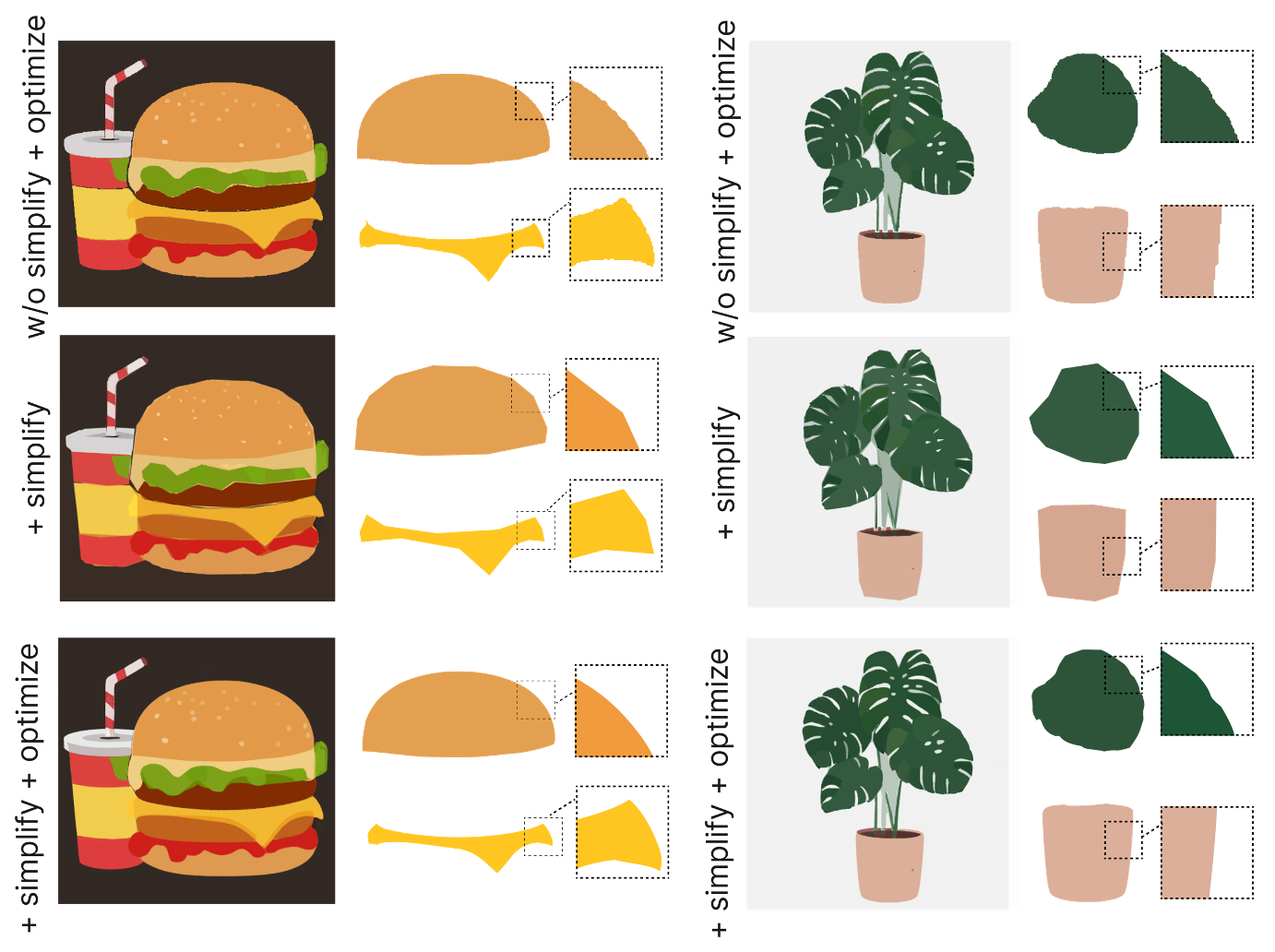}
    \caption{Comparison of structure-wise vectors with and without structure-wise vector optimization.}
  \label{fig:diff_struct_opt}
\end{figure}

\paragraph{Optimization of Structure-wise Vectors} We investigated the effectiveness of structure-wise vector optimization compared to direct initialization of structure-wise vectors from segmented masks, without optimization. Figure~\ref{fig:diff_struct_opt} shows with the optimization, the vector boundaries become neater and more refined.

\begin{figure}[!htb]
  \centering    \includegraphics[width=1.\linewidth]{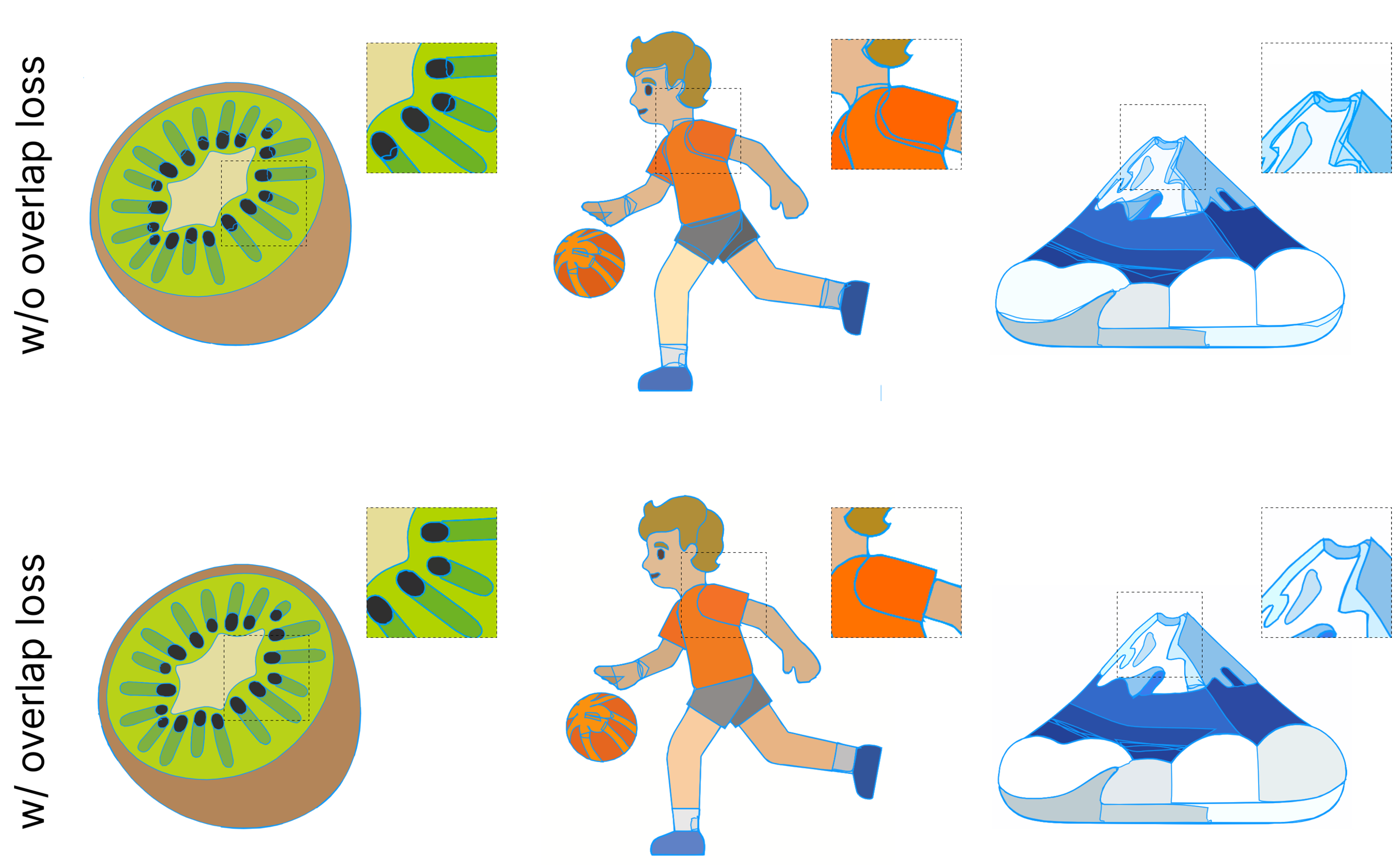}
    \caption{Comparison of structure-wise vectors with and without the \textit{overlap loss}. }
  \label{fig:diff_overlap_loss}
\end{figure}

\paragraph{Overlap Loss} We examined the impact of the overlap loss $\mathcal{L}_{\textit{overlap}}$ in structural construction. Figure~\ref{fig:diff_overlap_loss} shows with overlap loss, the vector boundaries are aligned better than those without the overlap loss. 

\section{Implementation Details}
\label{sec:details}
\paragraph{Initialization of Vector Primitives} Both structure-wise and visual-write vectors are initialized as closed shape of cubic Bézier curves.  For initialization, the boundary of the mask (for structure-wise vectors) or the boundary of the top-K connection area (for visual-wise vectors) is simplified using the Douglas–Peucker algorithm. This algorithm reduces the number of points while ensuring the simplified boundary remains within a distance $\epsilon$ ($\epsilon$ = 5.0 in our work) from the original boundary. 


\paragraph{Comparison Alignment} We compared our method with LIVE, DiffVG, O\&R, and SGLIVE, under the same number of vectors $N$ (64, 128, 256). For DiffVG, we initialized and optimized $N$ vectors. For O\&R, we initialized $4N$ vectors and optimized, then reduced the count to $N$. For LIVE and SGLIVE, the process involves adding vectors in blocks of increasing sizes, following an order of 8, 8 16, 32, 64, and 128. Vectors are added until the total number of vectors added equals $N$. Our method prioritized adding structure-wise primitives, up to a maximum of $N/2$, with the remaining count filled by visual-wise vectors to ensure the total reached $N$.


\section{Additional Results}

\begin{figure*}[!htb]
  \centering    \includegraphics[width=1.\linewidth]{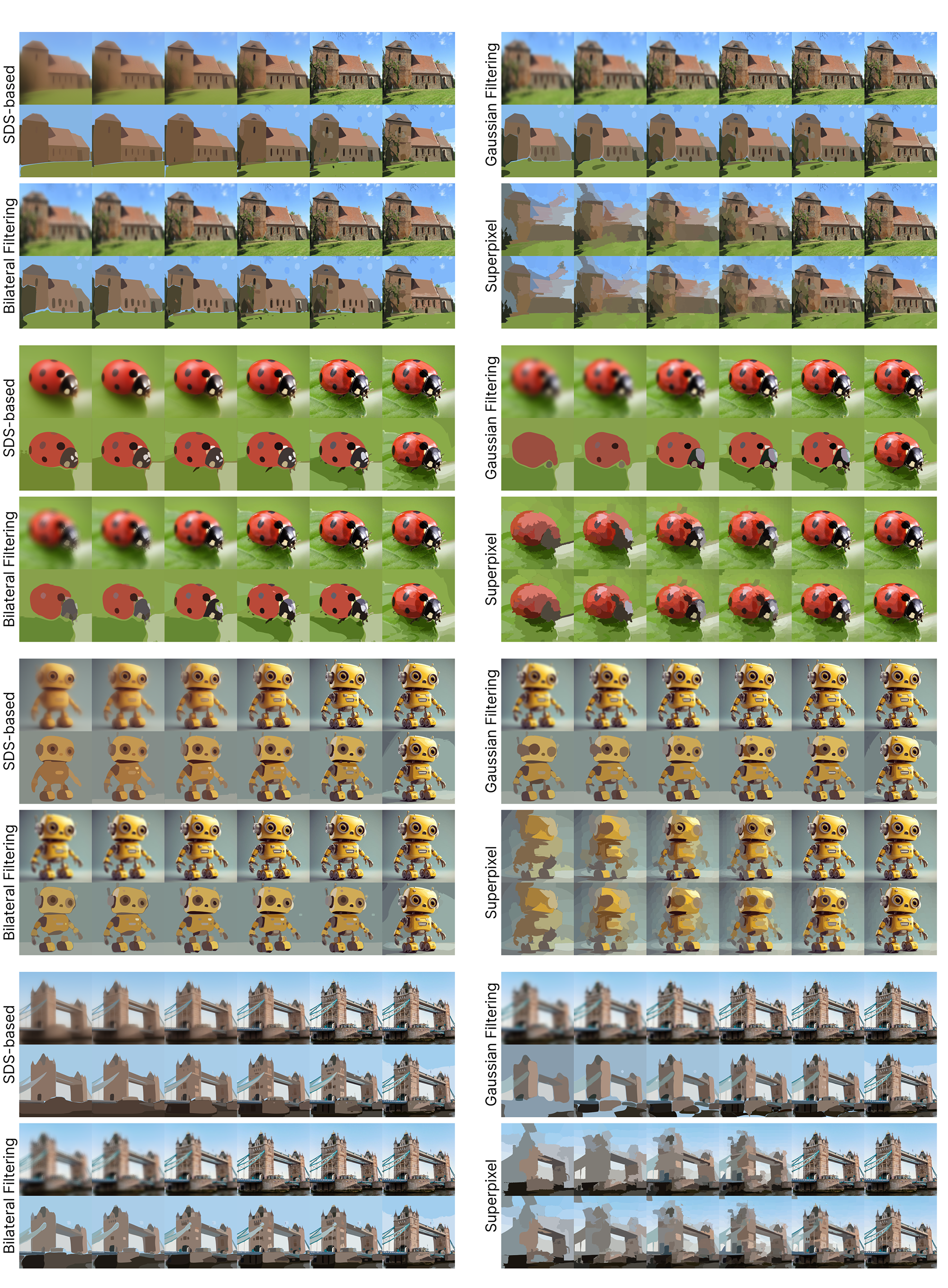}
    \caption{Comparison of vector layers between SDS-based method and three conventional image simplification methods}
  \label{fig:diff_sim_supp}
\end{figure*}

\paragraph{Comparison among Different Image Simplification Methods}Gaussian filtering generates four levels of simplified images by varying the kernel size of the Gaussian filter to 2, 6, 10, and 14. 
Bilateral filtering produces four levels of simplified images by setting the parameters (diameter, $\sigma_{Color}$, $\sigma_{Space}$) to $(10 + 5N, 100 + 50N, 100 + 50N)$, where $N = 0, 1, 2, 3$. Superpixel algorithm achieves four levels of simplification by reducing the number of superpixels the image is divided into, using values of 400, 200, 100, 50. Figure~\ref{fig:diff_sim_supp} presents additional results of reconstructed vector layers obtained with these different image simplification methods. 

\begin{figure*}[!htb]
  \centering    \includegraphics[width=1.\linewidth]{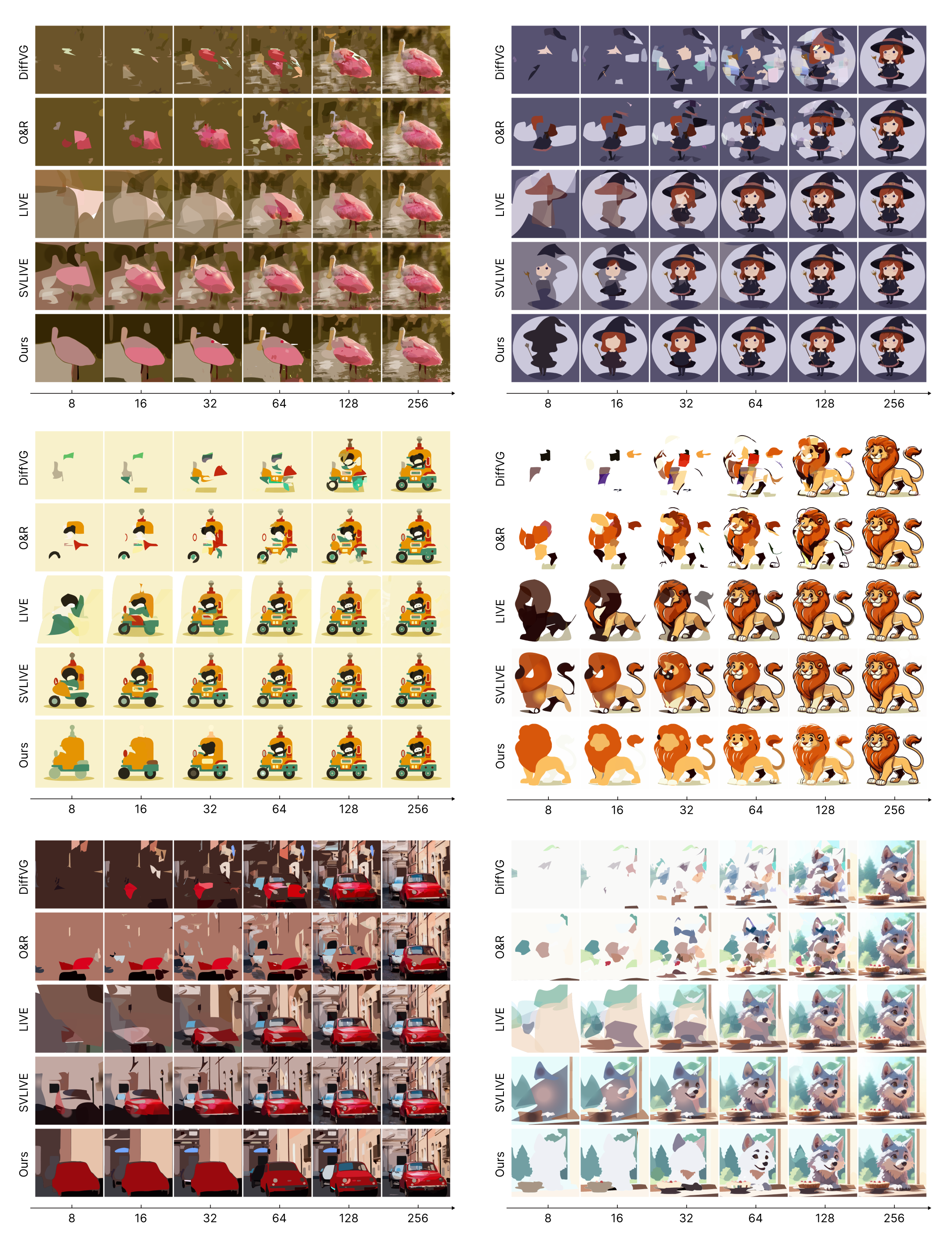}
    \caption{Comparison of vector layers between SDS-based method and three conventional image simplification methods}
  \label{fig:diff_sim_supp1}
\end{figure*}

\begin{figure*}[!htb]
  \centering    \includegraphics[width=1.\linewidth]{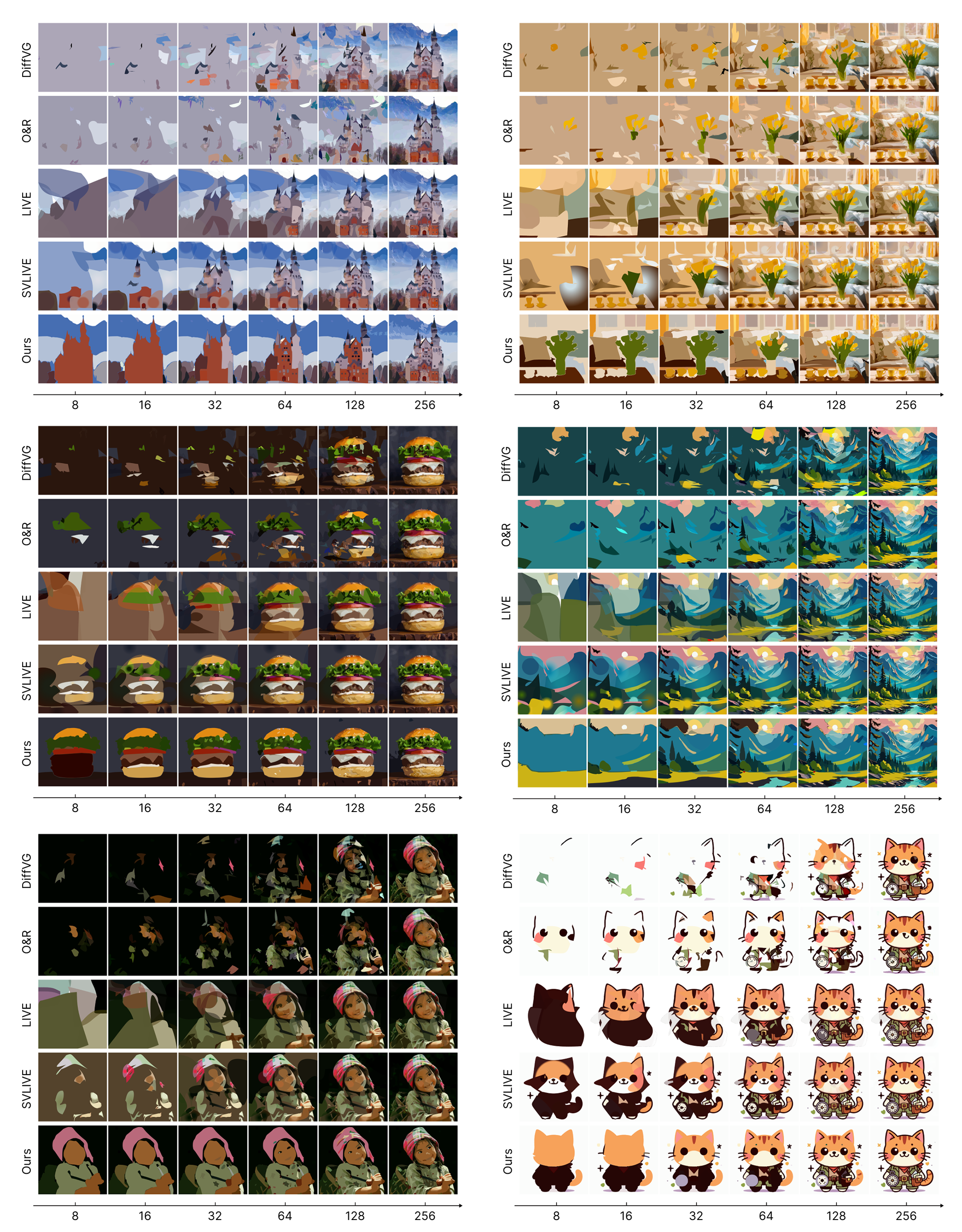}
    \caption{Comparison of vector layers between SDS-based method and three conventional image simplification methods}
  \label{fig:diff_sim_supp2}
\end{figure*}

\begin{figure*}[!htb]
  \centering    \includegraphics[width=1.\linewidth]{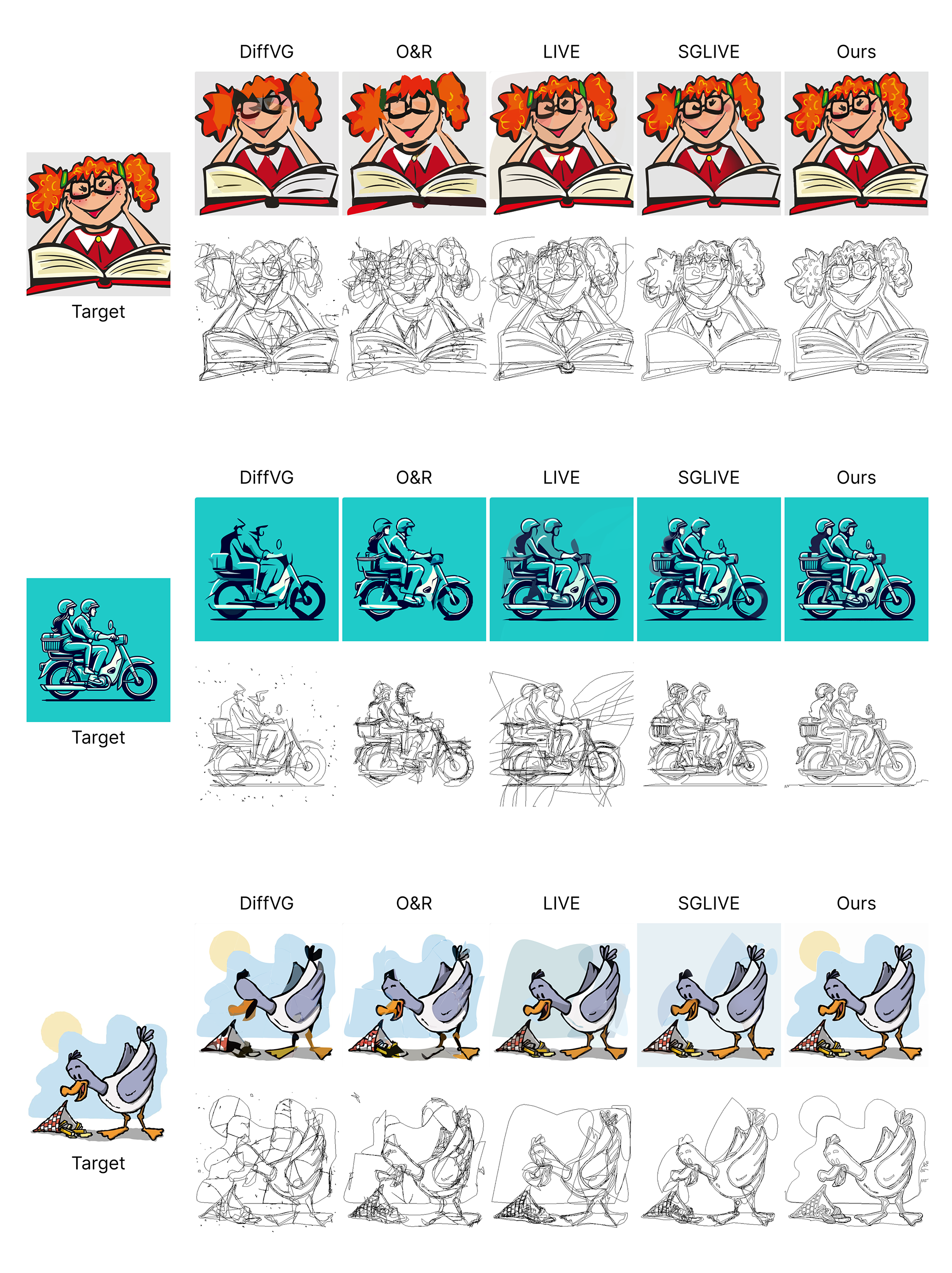}
    \caption{Comparison between our method and four state-of-the-art vectorization methods, vectorized with 128 primitives.}
  \label{fig:diff_visual1}
\end{figure*}

\begin{figure*}[!htb]
  \centering    \includegraphics[width=1.\linewidth]{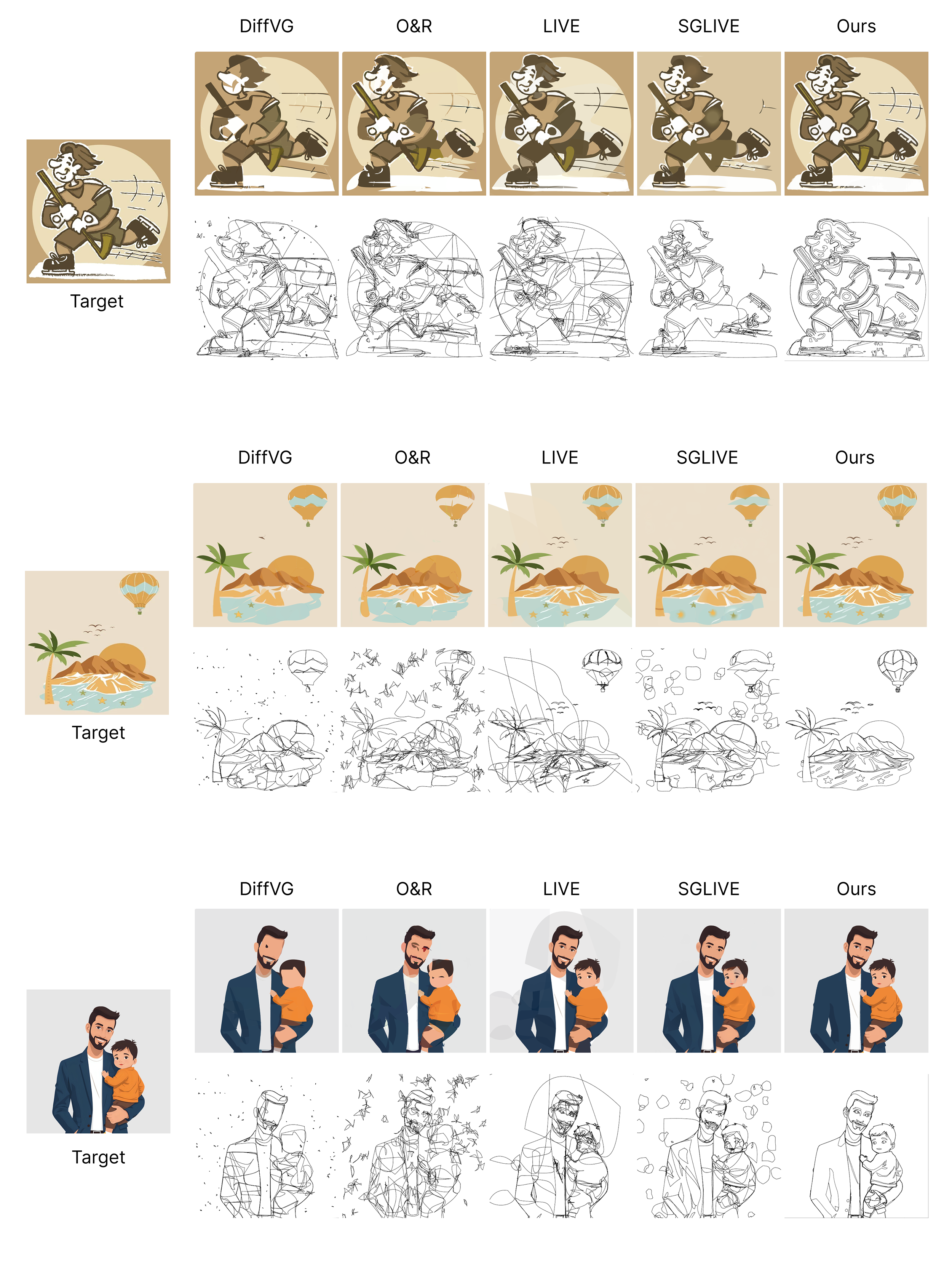}
    \caption{Comparison between our method and four state-of-the-art vectorization methods, vectorized with 128 primitives.}
  \label{fig:diff_visual2}
\end{figure*}

\begin{figure*}[!htb]
  \centering    \includegraphics[width=1.\linewidth]{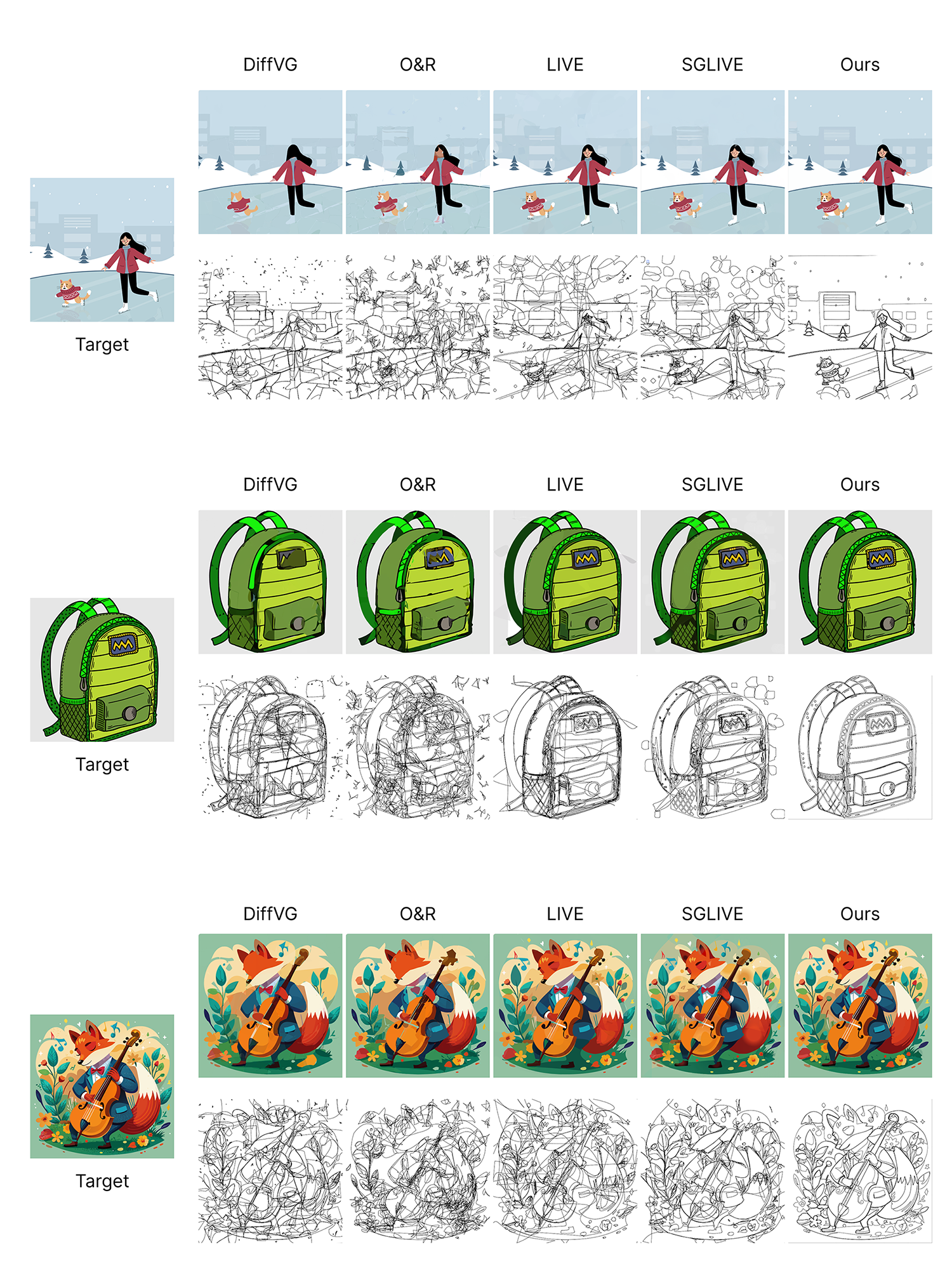}
    \caption{Comparison between our method and four state-of-the-art vectorization methods,  vectorized with 256 primitives.}
  \label{fig:diff_visual3}
\end{figure*}

\begin{figure*}[!htb]
  \centering    \includegraphics[width=1.\linewidth]{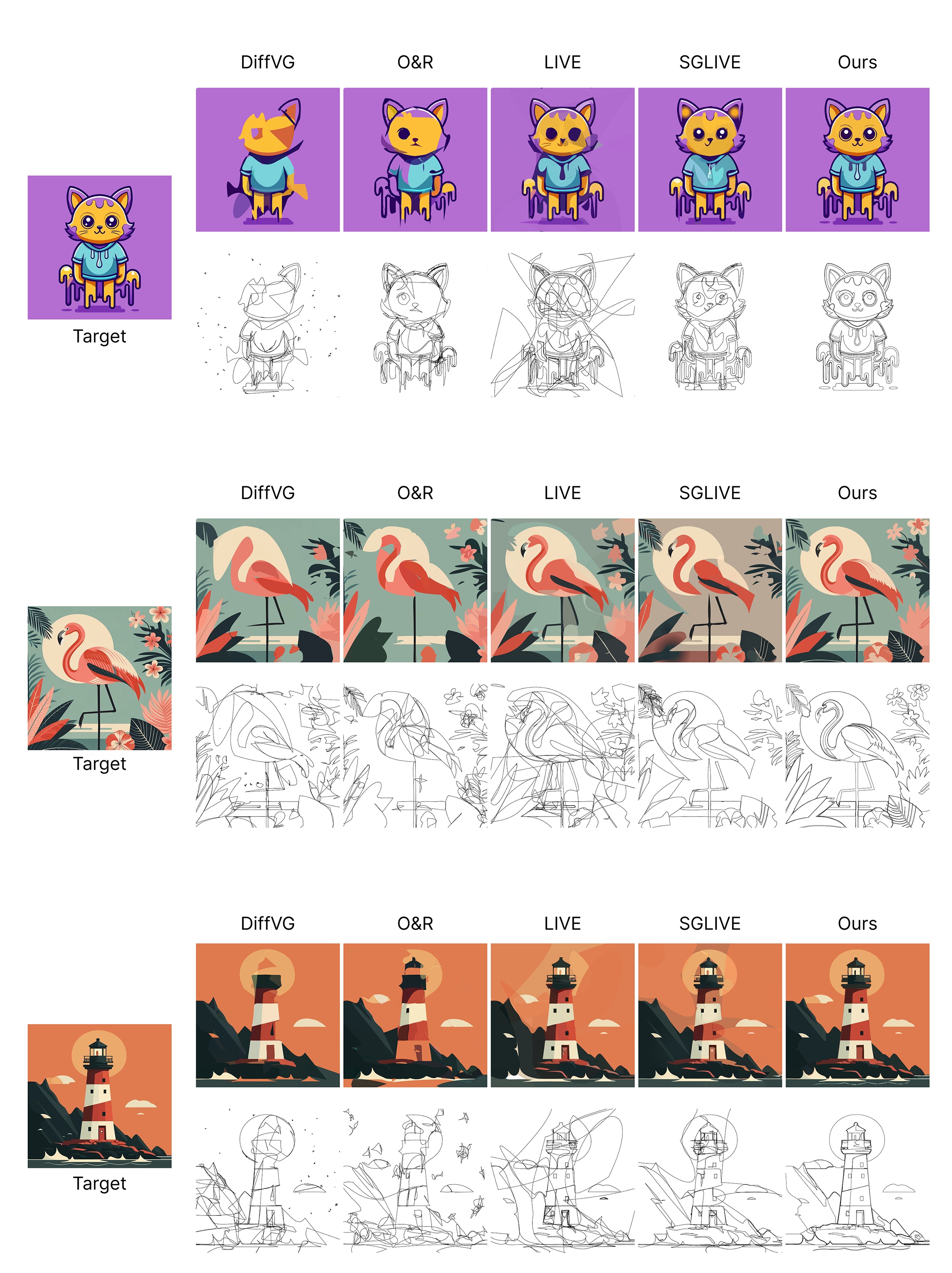}
    \caption{Comparison between our method and four state-of-the-art vectorization methods, vectorized with 64 primitives.}
  \label{fig:diff_visual4}
\end{figure*}

\paragraph{Comparison with Different Vectorization Methods} Figure~\ref{fig:diff_sim_supp1} and Figure~\ref{fig:diff_sim_supp2} shows additional results of vector layers constructed by our method and four methods. Figure~\ref{fig:diff_visual1}, \ref{fig:diff_visual2}, \ref{fig:diff_visual3} and \ref{fig:diff_visual4} compares the difference in visual fidelity and boundaries. 

\end{document}